\documentclass[journal,10pt,twocolumn,twoside]{IEEEtran}
\usepackage{mathtools}
\usepackage{empheq}
\usepackage{algpseudocode, algorithm, tabularx}
\usepackage{algorithmicx}
\usepackage{subfig}
\usepackage{varwidth}
\usepackage{url}
\usepackage{amssymb}
\usepackage{amsthm}
\usepackage{mathtools}
\usepackage{changes}
\usepackage{hyperref}
\usepackage{xcolor}
\usepackage[short]{optidef}
\usepackage{balance}

\DeclareMathOperator{\E}{\mathbb{E}}

\makeatletter
\newcommand{\multiline}[1]{%
	\begin{tabularx}{\dimexpr\linewidth-\ALG@thistlm}[t]{@{}X@{}}
		#1
	\end{tabularx}
}
\renewcommand{\ALG@name}{Alg.}
\makeatother

\ifCLASSOPTIONcompsoc
  \usepackage[nocompress]{cite}
\else
  \usepackage{cite}
\fi
\ifCLASSINFOpdf
\else
\fi
\begin{document}
%
\title{Sum-Rate Maximization for UAV-assisted Visible Light Communications using NOMA: Swarm Intelligence meets Machine Learning}

\author{Quoc-Viet Pham, 
	Thien Huynh-The, 
	Mamoun Alazab, 
	Jun Zhao, 
	and Won-Joo Hwang 
\IEEEcompsocitemizethanks{\IEEEcompsocthanksitem 
	Quoc-Viet Pham is with the Research Institute of Computer, Information
	and Communication, Pusan National University, Busan 46241, South Korea. 
	Thien Huynh-The is with the ICT Convergence Research Center, Kumoh National Institute of Technology, Gyeongsangbuk-do, 39177 Korea. 
	Mamoun Alazab is with the College of Engineering, IT and Environment, Charles Darwin University, Casuarina, NT 0810, Australia. 
	Jun Zhao is with the School of Computer Science and Engineering, Nanyang Technological University, 50 Nanyang Avenue, 639798 Singapore. 
	Won-Joo Hwang is with the Department of Biomedical Convergence
	Engineering, Pusan National University, Busan 46241, South Korea.
	Email: \{vietpq@pusan.ac.kr, 
	thienht@kumoh.ac.kr, 
	alazab.m@ieee.org, 
	junzhao@ntu.edu.sg, 
	wjhwang@pusan.ac.kr\}.}}

\markboth{Published in IEEE Internet of Things Journal}{Q.V. PHAM \MakeLowercase{\textit{et al.}}: Sum-Rate Maximization for UAV-assisted VLC using NOMA: Swarm Intelligence meets Machine Learning}


%


\IEEEtitleabstractindextext{%
\begin{abstract}
As the integration of unmanned aerial vehicles (UAVs) into visible light communications (VLC) can offer many benefits for massive-connectivity applications and services in 5G and beyond, this work considers a UAV-assisted VLC using non-orthogonal multiple-access. More specifically, we formulate a joint problem of power allocation and UAV's placement to maximize the sum rate of all users, subject to constraints on power allocation, quality of service of users, and UAV's position. Since the problem is non-convex and NP-hard in general, it is difficult to be solved optimally. Moreover, the problem is not easy to be solved by conventional approaches, e.g., coordinate descent algorithms, due to channel modeling in VLC. Therefore, we propose using harris hawks optimization (HHO) algorithm to solve the formulated problem and obtain an efficient solution. We then use the HHO algorithm together with artificial neural networks to propose a design which can be used in real-time applications and avoid falling into the ``local minima" trap in conventional trainers. Numerical results are provided to verify the effectiveness of the proposed algorithm and further demonstrate that the proposed algorithm/HHO trainer is superior to several alternative schemes and existing metaheuristic algorithms. 
\end{abstract}

}

%
\maketitle
\IEEEdisplaynontitleabstractindextext
\IEEEpeerreviewmaketitle

\section{Introduction}
\label{Sec:Introduction}
Visible light communications (VLC), unmanned aerial vehicles (UAV), and non-orthogonal multiple access (NOMA) are envisioned as three key technologies in 5G and beyond \cite{Marshoud2016NOMA_VLC, Ding2017aSurveyOnNOMA, Pham2019ASurvey_MEC}. On the one hand, VLC using
light-emitting diodes (LEDs) has emerged as a promising technology in optical wireless communication and recently received much attention from both industry and academia due to lots of offered advantages, such as easy installation, high data rate communication, low cost and power consumption, and high security \cite{Marshoud2016NOMA_VLC, Nguyen2010Matlab}. On the other hand, NOMA has been regarded as one of the key-enabler technologies in 5G networks and beyond. Unlike orthogonal multiple access (OMA) technologies, NOMA can serve many users using the same time-frequency resource. Thus NOMA is suitable for applications with massive connectivity, e.g., the Internet of Things (IoT) \cite{Pham2019CoalitionalGames, Ding2017aSurveyOnNOMA}. Besides VLC and NOMA, UAV has been recently exploited as a means to boost the capacity and coverage of wireless networks \cite{Pham2019ASurvey_MEC}. Two key features of UAV are high mobility and the provision of line-of-sight (LoS) connections.  

\subsection{Related Works}
\label{Subsec:RelatedWork}
Researches on VLC and NOMA have recently attracted significant attention. The first integration between NOMA and downlink VLC was studied in \cite{Marshoud2016NOMA_VLC} to enhance the achievable data rate. The authors discussed some reasons for a good interplay between NOMA and VLC. For instance, NOMA performs well at high signal-to-noise ratio (SNR), which can be guaranteed in VLC systems thanks to the short distance from the LED transmitter to users and LoS propagation. Moreover, the channels stay almost constant in VLC systems, which is vital to NOMA functionalities in performing successive interference cancellation (SIC) operation. Besides, the LED as a small access point can provide services to a few users, while NOMA typically multiplexes a few number of users in a cluster. Moreover, motivated by the fact that users with poor channel conditions should utilize higher transmit power due to severe inter-cluster interference, a gain ratio power allocation (GRPA) scheme was proposed in \cite{Marshoud2016NOMA_VLC}. The sum logarithmic data rate for NOMA-enabled downlink VLC networks was considered in \cite{Yang2017FairNOMA}. Power allocation schemes were investigated in \cite{Yin2016Performance, Zhang2017UserGrouping} to maximize the sum rate of all the users. These studies showed that VLC systems using NOMA can offer higher spectral efficiency compared with OMA counterparts. An experimental demonstration of VLC systems using optical power-domain NOMA was recently reported in \cite{Lin2019OpticalPD}.

The integration of NOMA and UAV has been recently studied as a key-enabler technology for massive connectivity and coverage enhancement in 5G and beyond. In \cite{Nasir2019UAV_Enabled}, a joint optimization problem of bandwidth allocation, power allocation, UAV altitude, and antenna beamwidth was formulated for NOMA-enabled UAV downlink networks, which was solved using a path-following algorithm. It was shown in \cite{Mu2019UplinkNOMA} that NOMA can be applied for uplink cellular networks to lower the UAV completion time while satisfying the quality of service (QoS) requirement of ground users (GUs). A NOMA random access scheme for uplink UAV-assisted wireless communications was studied in \cite{Seo2019UplinkNOMA}, where the proposed access algorithm is compared with the existing slotted ALOHA. Joint optimization of UAV's placement and power allocation was considered in \cite{Liu2019Placement} and \cite{Sohail2019EnergyEfficient} for the network sum rate and energy efficiency, respectively. NOMA precoding vectors at the BS and UAV were optimized for UAV-assisted cellular communications in \cite{Zhao2019JointTrajectory} and for cellular-connected UAV networks in \cite{Pang2019UplinkPrecoding}. 
Recently, the authors in \cite{Yang2019PowerEfficient} developed an iterative algorithm to minimize the total power consumption by optimizing user association and UAV location. 

{\color{black}To optimize UAV/NOMA/VLC systems, there have been some tools like network optimization \cite{Nasir2019UAV_Enabled, Zhao2019JointTrajectory, Yang2017FairNOMA}, game theory \cite{Pham2019CoalitionalGames}, and machine learning (ML) \cite{Wang2019DeepLearningUAV}. 
Especially, due to the emergence of new applications and technologies, ML has found many applications in 5G wireless and communication networks. For example, deep learning was considered in \cite{Wang2019DeepLearningUAV} for an extended version of the previous work \cite{Yang2019PowerEfficient}, federated learning was used in \cite{chen2019joint, yang2019energy} for several problems at the wireless edge, and deep reinforcement learning was comprehensively reviewed in \cite{luong2019applications}, where it can be used to solve many problems, e.g., wireless caching, edge computing, and network security. Motivated by a competitive performance with high reliability and fast convergence, swarm intelligence is considered as a key approach for optimizing the 5G and beyond networks. Harris Hawks Optimizer (HHO), as a swarm intelligence technique, is one of the most recent
algorithms that has been popular since the proposal \cite{Heidari2019HHO}. It has been considered in many engineering problems, e.g., the control chart patterns recognition for the manufacturing industry in \cite{golilarz2019new}, the design of microchannel heat sinks for the minimization of entropy generation in \cite{abbasi2019application}, and resource allocation in wireless networks \cite{Girmay2019JointChannel, Pham2020Whale}, which all show supervisor performance of the HHO algorithm. In the following parts of this paper, we will show that HHO is considered suitable for our problem and can achieve competitive performance compared with several state-of-the-art alternatives.}

\subsection{Motivation and Contributions}
\label{Subsec:Motivation}
To the best of our knowledge, this work is the first attempt to study the integration of UAV, NOMA, and VLC, which can offer many benefits compared with existing studies. First, a UAV-assisted VLC network with NOMA can provide illumination and communication services simultaneously for more users than VLC with conventional OMA schemes, thus realizing massive connectivity requirements for 5G and beyond. Second, UAVs using radio frequency (RF) resources for communications are energy-consuming, which is opposed to the fact that UAVs typically have a finite battery. This challenge can be addressed by powering UAVs through energy harvesting and wireless power transfer \cite{Huang2019Wireless}, and now can be further improved by equipping UAV with VLC capabilities. Next, UAVs possess the features of high mobility and flexibility to ensure LoS wireless links for VLCs. Therefore, the SIC operation in NOMA can be enhanced and the network performance increases accordingly. Finally, some concepts and testbed experiments have been conducted to verify the applications and practicability of UAVs in VLCs \cite{Deng2018Twinkle, Nurzhan2019Unmanned}. 

In this work, we consider a UAV-assisted VLC network using NOMA, where a UAV is equipped with a LED to provide both illumination and communications, and downlink transmissions between the UAV and GUs follow the NOMA principle. In summary, features and contributions offered by our work can be summarized as follows: 
\begin{itemize}
	\item We formulate an optimization problem to maximize the sum rate of all GUs subject to constraints on power allocation (e.g., budget at the UAV, peak optical intensity, and non-negative signals), QoS requirement of GUs, and UAV's placement. 
	
	\item Since the formulated optimization problem is non-convex and NP-hard in general, we propose using a swarm algorithm, namely HHO \cite{Heidari2019HHO}, to solve the problem. Compared with conventional methods such as gradient-based approaches, the HHO does not depend on gradient information of the objective functions and constraints, and is easy to implement because of its simplicity. Moreover, the HHO is capable of keeping a balance between exploration and exploitation phases, so it can avoid the possibility of being trapped in local optima and can be considered as a global optimizer.
	
	\item Moreover, common approaches, e.g., path-following procedure and coordinate descent method, are not applicable to solve the problem considered in this paper. Our proposed HHO-based algorithm finds solutions for UAV's placement and power allocation simultaneously, whereas the coordinate descent method needs to decompose the original problem into subproblems and solve them alternatively in an iterative manner. {\color{black}For example, \cite{Zhao2019JointTrajectory} proposed optimizing the joint trajectory and precoding problem by alternatively solving the user scheduling and trajectory optimization. Besides, successive convex approximation (SCA) technique is usually used to solve such subproblems (e.g., power allocation and UAV trajectory) \cite{Pang2019UplinkPrecoding, Nasir2019UAV_Enabled, Mu2019UplinkNOMA}.} However, the SCA technique is workable only if the channel gains follow an exponential-distance model, {\color{black}i.e., the channel gain is given as $ h = \psi d^{\alpha/2} $, where $ \psi $ is a constant, $ d $ is the distance, and $ \alpha $ is the path loss exponent. Therefore, this technique is not amenable to channel models in VLC, which will be presented in Section~\ref{Subsec:NetworkSetting}.} Our proposed HHO-based algorithm relaxes this assumption and is thus suitable for channel modeling in VLC. 
	
	\item For the first time, we adopt the HHO algorithm to optimize the connection weights and biases of artificial neural networks (ANNs). The inputs to the ANN are the locations of GUs, while the outputs are the UAV's placement and power allocation. Our proposed trainer is named HHOFNN, can avoid local solutions as in conventional algorithms (e.g., back propagation and conjugate gradient), and is able to achieve higher sum-rate than well-known metaheuristic optimizers: particle swarm optimization (PSO), evolution strategy (ES), and genetic algorithm (GA). 
	
	\item Through numerical simulation results, it is shown that the proposed HHO for power allocation and placement (HHOPAP) algorithm outperforms several existing algorithms under various simulation settings, including two NOMA schemes - gain ratio power allocation (GRPA) proposed in \cite{Marshoud2016NOMA_VLC} and random placement (RandP), and a conventional orthogonal frequency-division multiplexing access (OFDMA) scheme. 
\end{itemize}

The remaining of this paper is organized as follows. The system model and the problem formulation are presented in Section~\ref{Sec:Problem_Formulation}. The HHO-based algorithm is proposed in Section~\ref{Sec:HHO}. Our proposed trainer for ANNs is explained in Section~\ref{Sec:HHO_ANN}. In Section~\ref{Sec:Simulation}, numerical results are provided and discussed. Finally, conclusions are drawn in Section~\ref{Sec:Conclusion}.

\begin{figure}[t]
	\centering
	\includegraphics[width=1.00\linewidth]{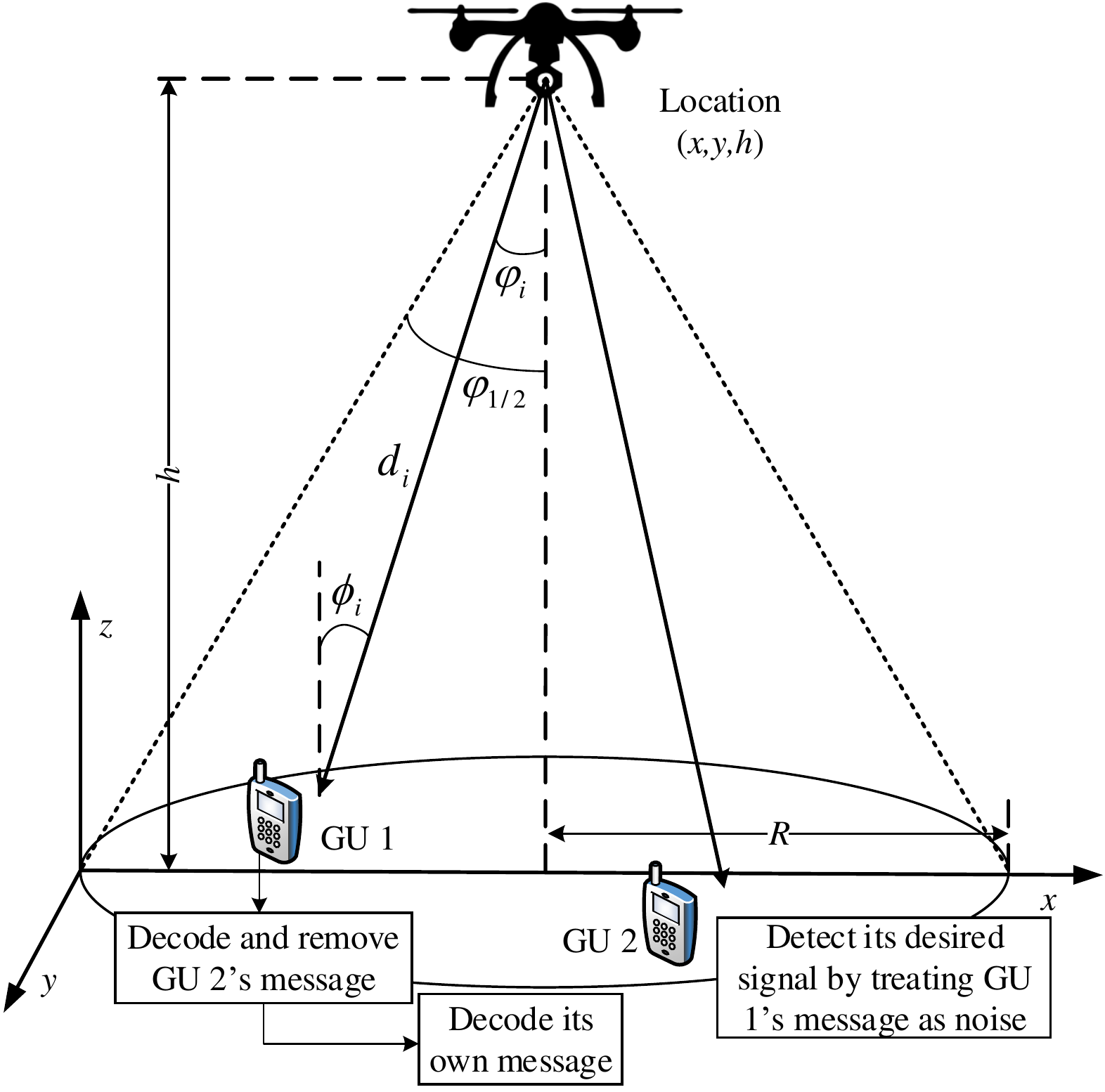}
	\caption{Illustration of a VLC network with UAV and NOMA.}
	\label{Illustration}	
\end{figure}

\section{Network Setting and Problem Formulation}
\label{Sec:Problem_Formulation}
\subsection{Network Setting}
\label{Subsec:NetworkSetting}
We consider a network setting as in Fig.~\ref{Illustration}, where the {\color{black}rotary-wing UAV}\footnote{{\color{black}Generally, UAVs can be classified as rotary-wing UAVs and fixed-wing UAVs. Rotary-wing UAVs have the ability to hover, whereas fixed-wing UAVs need to move forward to remain aloft. Extension of the current work to consider rotary-wing UAVs is interesting and will be considered in our work.}} is equipped with one LED transmitter to provide illumination and communication for $ N $ randomly positioned GUs under its coverage area. Denote by $ \mathcal{N}=\{1,\dots,N\} $ the set of GUs and by $ h_{i} $ the channel gain between the UAV and GU $ i $. We assume that the UAV-mounted LED transmitter simultaneously transmits the data to $ N $ GUs using the same time-frequency resource by appropriately adjusting its transmit power (i.e., power-domain NOMA). GUs apply the successive interference cancellation (SIC) technique to remove the messages transmitted for other GUs. In particular, in downlink NOMA, the decoding order follows the increasing order of the channel gains. Without loss of generality, we assume that $ h_{1} < \dots < h_{N} $, i.e., the first and last users are considered as the weakest and strongest GUs, respectively. With this assumption, GU $ i $ first decodes the message transmitted for GU $ 1 $ and subtracts it from the received signal, until GU $ (i-1) $.

For the channel model, we consider the LoS propagation path and use the generalized Lambertian emission model, where the channel direct current (DC) gain is proportional to the inverse of the squared distance between the LED transmitter and user \cite[Eq.~(10)]{Kahn1997WirelessIC}. Specifically, the DC channel gain of GU $ i $ is expressed by
\begin{equation}\label{Eq:ChannelGain}
h_{i} = \frac{A_{i}}{d_{i}^{2}} R_{0}(\varphi_{i}) T_{s}(\phi_{i}) g(\phi_{i}) \cos(\phi_{i})
\end{equation}
for $ 0 \leq \phi_{i} \leq \Phi_{i} $ and $ h_{i} = 0 $ otherwise, where $ \Phi_{i} $ is the field of view (FoV) at GU $ i $, $ \varphi_{i} $ and $ \phi_{i} $ are the angles of irradiance and incidence, respectively. In \eqref{Eq:ChannelGain}, $ A_{i} $ is the detection area, $ d_{i} $ the distance between the
UAV/LED transmitter and GU $ i $, $ T_{s}(\phi_{i}) $ is the gain of the optical filter, $ g(\phi_{i}) $ denotes the concentrator gain and is given by 
\begin{equation}\label{Eq:concentratorGain}
g(\phi_{i}) = \frac{n^{2}}{\sin^{2}\Phi_{i}}, \; 0 \leq \phi_{i} \leq \Phi_{i},
\end{equation}
with $ n $ being the internal refractive index. Also, $ R_{0}(\varphi_{i}) $ is the Lambertian radiant intensity and is given as 
\begin{equation}\label{key}
R_{0}(\varphi_{i}) = \frac{m+1}{2\pi} \cos^{m}\varphi_{i}
\end{equation}
where $ m $ is related to the transmitter semiangle at half power and 
$ m = -\ln 2 / \ln\left(\cos \varphi_{1/2}\right) $. 
In this paper, we fix the value of $ \varphi_{1/2} $ to $ 60^{\circ} $, i.e., $ m = 1 $.

We assume that the UAV, flying within a disc with the radius of $ R $, is fixed at the altitude $ h $. As we know from the literature on VLC, VLC can be used for location prediction, so we can assume that the UAV knows perfectly the locations of GUs. All GUs are positioned at the ground (their altitude is zero). The distance between the UAV and GU $ i $ is given as 
\begin{equation}\label{key}
d_{i} = \sqrt{\left( x_{u} - x_{i} \right)^{2} + \left( x_{u} - x_{i} \right)^{2} + h^{2}},
\end{equation}
where $ (x_{u},y_{u},h) $ and $ (x_{i},y_{i},0) $ are the coordinate vectors of the UAV and GU $ i $, respectively. We note that the values of the angles of irradiance and incidence can be achieved from $ \cos \varphi_{i} = \cos \phi_{i} = h/d_{i}  $.

Denote by $ p_{i} $ and $ s_{i} $ the transmit allocated power and
the message for GU $ i $, respectively. The composite signal at the UAV can be expressed as follows:
\begin{equation}\label{key}
s = \sum\nolimits_{i \in \mathcal{N}}\sqrt{p_{i}}s_{i} + A,
\end{equation}
where $ A $ is a DC offset/bias added to ensure the positive instantaneous intensity of the transmitted signal. To maintain non-negativity of the transmitted signal $ s $, the
following constraint has to be satisfied:
\begin{equation}\label{key}
\sum\limits_{i \in \mathcal{N}}\sqrt{p_{i}} \leq \frac{A}{\delta},
\end{equation}
where $ \delta $ is a coefficient, which is determined by modulation
order of the pulse amplitude modulation (PAM) employed in
the network. Power allocation of the UAV should be limited due to its power budget and eye safety \cite{Kahn1997WirelessIC}. To consider eye safety, the transmitted optical intensity should be limited by a peak
optical intensity $ B $ \cite{Yang2017FairNOMA}, as the following constraint
\begin{equation}\label{Eq:achievableRate}
\sum\limits_{i \in \mathcal{N}}\sqrt{p_{i}} \leq \frac{B-A}{\delta},
\end{equation}

According to the downlink NOMA principle, the achievable data rate of GU $ i $ can be given as follows \cite{Ding2017aSurveyOnNOMA, Pham2018AlphaFairness}:
\begin{equation}\label{key}
R_{i} = \log_{2}\left(1 + \frac{h_{i}p_{i}}{n_{0} + \sum\limits_{j \in \mathcal{N} : h_{j} > h_{i}}h_{i}p_{j}}\right),
\end{equation}
where $ p_{i} $ is the transmit power allocated to GU $ i $, $ h_{i} $ is the channel gain between the UAV and GU $ i $, and $ n_{0} $ is the noise power at GUs. The NOMA rate in~\eqref{Eq:achievableRate} is attained through implementations of the superposition of signals at the transmitter side (i.e., BS) and successive interference cancellation (SIC) at the receiver side (i.e., GUs). To perform SIC operation at GUs successfully, the transmit power for each GU should be assigned properly, which needs to satisfy the following constraints \cite{Ali2016Dynamic}:
\begin{equation}\label{Eq:SIC_Constraint}
p_{i}\bar{h}_{i+1} - \sum\nolimits_{j = i+1}^{N}p_{j}\bar{h}_{i+1} \geq \theta, \; i = 1, \cdots, N-1,
\end{equation}
where $ \bar{h}_{i} = h_{i}/n_{0}, \; \forall i \in \mathcal{N} $ and $ \theta $ is the minimum power difference required to distinguish between the signal to be decoded and the remaining non-decoded signals.

\subsection{Problem Formulation}
\label{Subsec:ProblemFormulation}
With the objective of maximizing the sum rate of all GUs\footnote{Note that we do not limit this study to a specific objective, which means other objective functions can be used instead of the sum rate in this paper. Moreover, fairness among GUs can be studied by considering a well-known family of $ \alpha $-utility functions that has been discussed in \cite{Pham2017Fairness}. }, the optimization problem can be formulated as follows:
\begin{subequations}
	\label{P1}
	\begin{align}
	& \underset{\{\boldsymbol{w},\boldsymbol{p}\}}{\max}
	& & \sum\limits_{i \in \mathcal{N}}\log_{2}\left(1 + \frac{h_{i}p_{i}}{n_{0} + \sum\limits_{j \in \mathcal{N} : h_{j} > h_{i}}h_{i}p_{j}}\right) \label{P1:a} \\
	& \text{s.t.}
	& & p_{i} \geq 0, \forall i \in \mathcal{N}, \label{P1:b} \\
	&&& \sum\nolimits_{i \in \mathcal{N}}p_{i} \leq P_{\max}, \label{P1:c} \\
	&&& \sum\nolimits_{i \in \mathcal{N}}\sqrt{p_{i}} \leq C, \label{P1:d} \\
	&&& p_{i}\bar{h}_{i+1} - \sum\limits_{j = i+1}^{N}p_{j}\bar{h}_{i+1} \geq \theta, \; i = 1, \cdots, N-1, \label{P1:e}\\
	&&& R_{i} \geq R_{i}^{\text{th}}, \; \forall i \in \mathcal{N}, \label{P1:f} \\
	&&& x_{u}^{2} + y_{u}^{2} \leq R^{2}, \label{P1:g}
	\end{align}
\end{subequations} 
where $ \boldsymbol{w} = \{x_{u},y_{u}\} $ (placement vector), $ \boldsymbol{p} = \{p_{1}, \dots, p_{N}\} $ (power allocation), and $ C = \delta^{-1}\min\{A, B-A\} $. The optimization variables $ (\boldsymbol{w},\boldsymbol{p}) $ are subject to several constraints. First, the transmit power is allocated to GUs on condition that the UAV power limitation and eye safety are guaranteed, as illustrated in \eqref{P1:b}-\eqref{P1:e}. Next, the QoS requirements of GUs are imposed in \eqref{P1:e}. Finally, the constraint \eqref{P1:f} indicates that the UAV should be deployed within a disc with the radius of $ R $. 

As the sum-rate maximization problem is non-convex NP-hard in general \cite{Pham2017Fairness}, the problem~\eqref{P1} is difficult to be solved optimally. Moreover, the channel model in VLC systems, illustrated in~\eqref{Eq:ChannelGain}, makes conventional convex approximation techniques not a proper tool to solve the problem~\eqref{P1}. In the next section, we will present the fundamentals of the HHO and utilize it as an optimizer of the problem~\eqref{P1}.

\section{Harris Hawks Optimization}
\label{Sec:HHO}

The HHO is a population-based metaheuristic algorithm, which was proposed by Heidari \cite{Heidari2019HHO} in 2019 based on the cooperative behaviors Harris' Hawks in hunting escaping preys, e.g., rabbits, mice, voles, and squirrels. Originally, the HHO can be applied to any continuous and unconstrained optimization problem. Mathematically, the hunting behavior of Harris' hawks can be modeled in three phases: 1) exploration, 2) exploitation, and 3) transition between exploitation and exploration. However, since there are several constraints, the problem~\eqref{P1} should be adjusted accordingly. Thus, we will also present a constraint-handling technique, which is to transform the problem~\eqref{P1} into a proper form.


\subsection{Exploration phase}
\label{Subsec:Exploration}
In this phase, Harris' hawks wait, observe, and minor the prey by their powerful eyes. If the hawks detect the prey, they can use two strategies to attack. In the first strategy, each hawk selects another random one from the family and decides the location to peach accordingly, whereas in the second strategy, the hawks perch on some high positions based on the cooperation with all the other family members. These strategies are considered with equal probability. In other words, if we consider a random $ q $ uniformly distributed in $ [0, 1] $, the first and second strategies are selected under the conditions $ q < 0.5 $ and  $ q \geq 0.5 $, respectively. Mathematically, these strategies are expressed by the following system.
\begin{align} 
& X(t+1) = X_{r}(t)-r_{1}\left | X_{r}(t)-2r_{2}X(t) \right |, \; \text{if } q \geq 0.5, \label{Ex:1stStrategy}\\
& X(t+1) = (X_{p}(t)-X_{a}(t))-r_{3}(LB+r_{4}(UB-LB)), \notag\\
& \phantom{cccccccccccccccccccccccccccccccccccccccccc} \text{if } q < 0.5, \label{Ex:2ndStrategy}
\end{align}
where $X(t)$ is the position vector of the hawks in the iteration $ t $ (i.e., $ X(t+1) $ is the position vector in the next iteration) and $ X_{p}(t) $ is the position of the prey, which is considered as the solution in the iteration $ t $. Here, $r_{1}$, $r_{2}$, $r_{3}$, $r_{4}$, and $q$ are numbers in the range $ [0, 1] $ that are created randomly in each iteration, $ LB $ and $ UB $ denote the upper and lower bounds of variables, respectively. 
In our problem, $ UB $ and $ LB $ of power allocation are set to be $ P_{\max} $ and $ 0 $, respectively. Similarly, those are $ R $ and $ 0 $ for the UAV's coordinate. 

In~\eqref{Ex:2ndStrategy}, $X_{r}(t)$ is the position vector of a hawk selected randomly from the current population and $X_{a}$ is the average position of the current population of hawks, which is given as follows: 
\begin{equation} \label{Eq:averageposition}
X_{a}(t) = \frac{1}{N}\sum_{i=1}^{N}X_{i}(t),
\end{equation}
where $ X_{i}(t) $ is the location of the $ i $-th hawk in the iteration $t$ and $ N $ is the population dimensionality. Thanks to the use of different random coefficients (in each iteration), the solution space can be diversified and thus the HHO algorithm can avoid the trap into local optima.

\subsection{Transition from exploration to exploitation}
To model the transition from exploration to exploitation, the energy of the prey is given as
\begin{equation} \label{Eq:energyPrey}
E = 2E_{0}(1-\frac{t}{T}),
\end{equation}
where $ t $ denotes the iteration index, $ T $ is the maximum number of iterations, and $ E_{0} $ is the initial energy of the prey, which is created randomly inside the range $ [-1 \; 1] $ in each iteration. When $ E_{0} $ decreases from $ 0 $ to $ -1 $, the prey is physically weakening; otherwise, the prey is strengthening. 

\begin{figure}[t]
	\centering
	\includegraphics[width=0.90\linewidth]{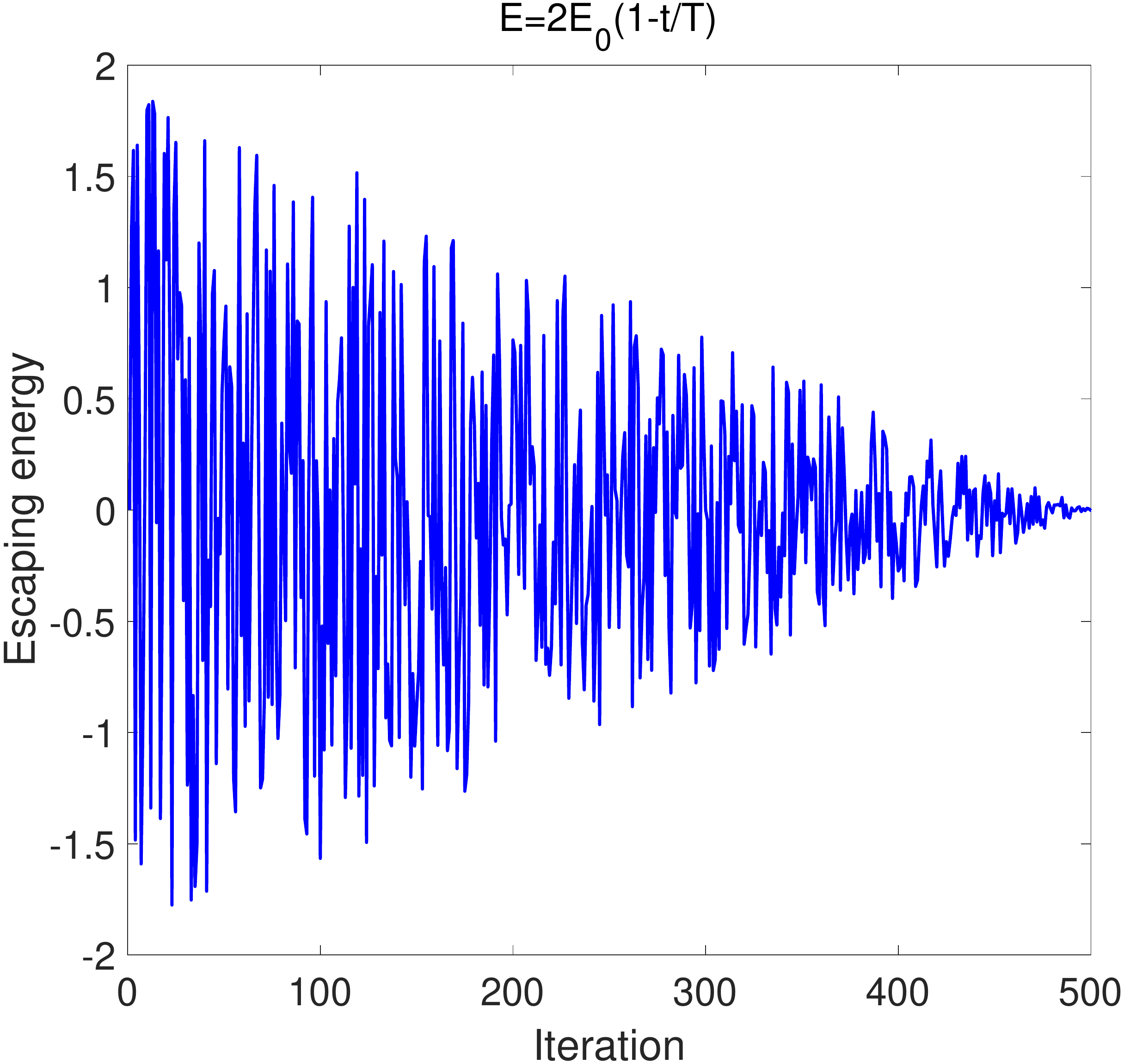}
	\caption{Evolution of the escaping energy over 500 iterations.}
	\label{Fig:E_Evolution}	
\end{figure}

When the value of $E_{0}$ decreases from $ 0 $ to $ -1 $, the rabbit is physically flagging, whilst when the value of $E_{0}$ increases from 0 to 1, it means that the rabbit is strengthening. Clearly, the value of $ E $ tends to decrease over the course of iterations. When $ |E| \geq 1 $, the exploration phase gets started, i.e., the algorithm tries to explore different regions in the solution space, whilst the exploitation phase is carried out when $ |E| < 1 $, i.e., the HHO algorithm exploits the neighborhood of solutions to update the position vectors for the next iteration. {\color{black}An example of the escaping energy $ E $ is showed in Fig.~\ref{Fig:E_Evolution}, where the maximum number of iterations $ T $ is set to be $ 500 $.}

%
%
%
%

\subsection{Exploitation phase}

In this phase, four attacking strategies are modeled according to the escaping probability of the prey and the chasing patterns of the hawks. Here, we define $ r $ as the escaping probability of the prey, where $ r \geq 0.5 $ and $ r < 0.5 $ indicate that the prey escapes from the attack unsuccessfully and successfully, respectively. Depending on the retained energy $ E $, the algorithm decides to besiege either hard or softly. Particularly, when $ |E| \geq 0.5$, the soft siege occurs, whereas when $|E| < 0.5$, the hard besiege happens.

\subsubsection{Soft besiege} This tactic takes place when $r\geq0.5$ and $|E|\geq0.5$, i.e.,  the prey has enough energy to escape but finally it fails to do, and the hawks encircle the prey softly to make it exhausted. This process is stimulated as follows:
\begin{align} 
& X(t+1) = \Delta X(t)-E\left |JX_{p}(t) - X(t)\right |, \label{Eq:SoftSiege} \\	
& \Delta X(t) = X_{p}(t) - X(t). \label{Eq:SoftSiege_deltaX} 
\end{align}
Here, $\Delta X(t)$ represents the difference between the position vectors of the prey and hawks in the iteration $ t $, $ J = 2(1-r_{5})$ indicates the jump strength the prey, and $ r_{5} $ is a random number inside the range $ [0, 1] $. Similar to random coefficients mentioned earlier, the value of $ r_{5} $ (and $ J $) is updated in each iteration to imitate the sudden movement of the prey.

\subsubsection{Hard besiege} This hunting tactic happens when $ r \geq 0.5$ and $|E| < 0.5$, , i.e., the prey does not have sufficient energy to escape and the hawks performs a hard siege before a surprise pounce (also known as seven kills, i.e., several hawks come from different directions and attack the prey \cite{dugatkin1997cooperation}). In this tactic, the position vector of the hawks is updated as
\begin{equation} \label{Eq:HardSiege}
X(t+1) = X_{p}(t) - E \left |\Delta X(t) \right |,
\end{equation}
where $ \Delta X(t) $ is given in Eq.~\eqref{Eq:SoftSiege_deltaX}.


\subsubsection{Soft besiege with progressive rapid dives} This tactic happens when $ |E| \geq 0.5$ and $ r < 0.5$, i.e., the prey has enough energy and escapes successfully, and the hawks softly besiege the prey to finally perform the surprise pounce. Inspired by the behavior of Harris's hawks in nature, it is supposed that the hawks can progressively select the best possible dive when they want to grab the prey. In this case, the next move of the hawks can be expressed as follows:
\begin{equation} \label{Eq:SoftSiege_progressive_nm}
Y = X_{p}(t) - E\left |JX_{p}(t) - X(t)\right |.
\end{equation}
Additionally, The hawks perform a corresponding dive when they observe a deceptive movement from the prey. To model such leapfrog movements, the L\'evy flight (LF) concept is utilized in the HHO algorithm. An LF is a special type of arbitrary walk, where the step lengths follow a heavy-tailed probability distribution \cite{Ling2017LevyFlight}. Specifically, the LF-based patterns can be attained by the following equation
\begin{equation} \label{Eq:LF_Movement}
Z = Y + S\times LF(D),
\end{equation}
where $ D $ is the population dimensionality and $ S $ is a random vector of size $ D $, and $ LF(\cdot) $ denotes the LF function. According to \cite{Heidari2019HHO, Xie2019ImprovedBHO}, the LF walk is modeled as follows:
\begin{equation} \label{Eq:LF} 
LF(x)=0.01 u \sigma / \left | v \right |^{1/\beta},
\end{equation}
where $u$ and $v$ are random values inside the range $ [0, 1] $, $\beta$ is generally set to be $ 1.5 $, and $ \sigma $ is given as
\begin{equation}\label{Eq:LF_sigma}
\sigma =\left ( \frac{\Gamma (1+\beta ) \sin(\pi\beta/2)}{\Gamma((1+\beta)/2) \beta\times2^{(\beta-1)/2})}  \right )^{\frac{1}{\beta}}.
\end{equation} 
Here, $ \Gamma(\cdot) $ denotes the standard \emph{Gamma} function and is expressed as $ \Gamma(z) = \int_{0}^{\infty} t^{z-1}e^{t}dt $.

The final step of this strategy is updating the position vectors of the hawks, which follows the rule below
\begin{equation} 
\label{Eq:SoftSiege_progressive_pos}
X(t+1)=
\begin{cases}
Y, & \text{if } F(Y) > F(X(t)), \\ 
Z, & \text{if } F(Z) > F(X(t)), \\
\end{cases}
\end{equation}
where $ F(\cdot) $ represents the fitness function. In the context of our problem, the fitness value is the total sum rate of all GUs plus the penalty imposed by the constraints, which will be shown in the following subsection.

\subsubsection{Hard besiege with progressive rapid dives} This hunting tactic happens when $ |E| < 0.5$ and $ r < 0.5$. In this case, the prey does not have enough energy to escape and the hawks carry out a hard siege to finally kill the prey. This tactic is similar to the soft besiege with progressive rapid dives except that the next move of the hawks is evaluated using the following rule
\begin{equation} \label{Eq:HardSiege_progressive_nm}
Y = X_{p}(t) - E\left |JX_{p}(t) - X_{a}(t)\right |,
\end{equation}
where the average position vector $ X_{a} $ is calculated according to Eq.~\eqref{Eq:averageposition}. Different from~\eqref{Eq:SoftSiege_progressive_nm}, where the hawks try to reduce the distance between the position of the prey and their current locations, in~\eqref{Eq:HardSiege_progressive_nm} the hawks attempt to decrease the distance between the position of the prey and the average location of Harris' hawks. Hence, the LF-based patterns and the next positions are updated using~\eqref{Eq:LF_Movement} and~\eqref{Eq:SoftSiege_progressive_pos}, respectively.

In summary, details of the HHO algorithm are summarized in Alg.~\ref{Alg:HHO}. As a population-based approach, a number of solutions evolve during the optimization process. In the experiment, we set $ S = 30 $, i.e., the population size is $ 30 $. Among solutions in the current population, the one that achieves the best fitness value is considered as the best solution, i.e., the position of the prey $ X_{p} $ (line~\ref{Alg:bestPosition}). Either of two phases, i.e., exploration and exploitation, is executed in an iteration, and the algorithm runs at most $ T $ iterations (line~\ref{Alg:T_iterations}). Using numerical results, we will show that the algorithm can converge to a final solution after a few tens of iterations. {\color{black}Moreover, thanks to a good balance between exploration and exploitation capabilities, the HHO algorithm can be regarded as an efficient global optimizer.}

\subsection{Mechanism to handle constraints}
The HHO in its original form was proposed for meta-heuristically solving unconstrained optimization problems. Since there are several constraints in our problem formulated in~\eqref{P1}, a natural question is how to incorporate these constraints. In fact, there are mechanisms to deal with constraints \cite{Jordehi2015Review}, e.g., penalty method, decoders, feasibility rules, stochastic ranking, $ \epsilon $-constrained method, multi-objective approach, and ensemble of constraint-handling techniques. As the most common approach in the evolutionary algorithm community, the penalty method is presented and used in this work. For more details of other methods, we invite interested readers to read \cite{Coello2002Theoretical, Yang2014Nature_Inspired}, and references therein.

The main idea of the penalty method is to define a fitness function so that the constrained problem~\eqref{P1} is transformed into an unconstrained problem. In particular, the penalty function can be expressed as follows:
\begin{equation}\label{Eq:fitnessFunction}
F(\boldsymbol{X}) = \sum\limits_{i \in \mathcal{N}}\log_{2}\left(1 + \frac{h_{i}p_{i}}{n_{0} + \sum\limits_{j \in \mathcal{N},j \neq i}h_{i}p_{j}}\right) + P(\boldsymbol{X}),
\end{equation}
where $ P(\cdot) $ is the penalty and $ \boldsymbol{X} = \{x_{u}, y_{u}, p_{1}, p_{2}, \dots, p_{N}\} $ is the optimization vector, which has the size of $ (2 + N) $. The penalty term can be calculated as follows:
\begin{align}\label{Eq:penaltyTerm}
P(\boldsymbol{X}) = 
& -\mu_{1} \left(\sum\limits_{i \in \mathcal{N}}p_{i} - P_{\max}\right)^{2} H\left(\sum\limits_{i \in \mathcal{N}}p_{i} - P_{\max}\right) \notag\\
& - \mu_{2} \left(\sum\limits_{i \in \mathcal{N}}\sqrt{p_{i}} - C\right)^{2} H\left(\sum\limits_{i \in \mathcal{N}}\sqrt{p_{i}} - C\right) \notag\\
& - \sum\limits_{i = 1}^{N-1}\mu_{i+2} \left( \theta - p_{i}\bar{h}_{i+1} + \sum\limits_{j = i+1}^{N}p_{j}\bar{h}_{i+1} \right)^{2} \notag\\
& \;\;\; \times H\left( \theta - p_{i}\bar{h}_{i+1} + \sum\limits_{j = i+1}^{N}p_{j}\bar{h}_{i+1} \right) \notag\\
& - \sum\limits_{i = 1}^{N}\mu_{i+N+1} \left(R_{i}^{\text{th}} - R_{i}\right)^{2} H\left(R_{i}^{\text{th}} - R_{i}\right) \notag\\
& - \mu_{2N+2} \left(x_{u}^{2} + y_{u}^{2} - R^{2}\right)^{2} H\left(x_{u}^{2} + y_{u}^{2} - R^{2}\right),
\end{align}
where $ \mu_{j}, \; j = \{1,\dots,2N+2\} $ are position constraints called penalty factors. Here, $ H(f_{j}(x)) $ denotes the indicator function of $ f_{j}(x), \; j = \{1,\dots,2N+2\} $. More specifically, $ H(f_{j}(x)) = 0 $ if $ f_{j}(x) \leq 0 $, whereas $ H(f_{j}(x)) = 1 $ if $ f_{j}(x) > 0 $. The minus sign in the penalty term~\eqref{Eq:penaltyTerm} is used since we consider maximization of the total sum rate. If the problem~\eqref{P1} is modeled as a minimization problem, the positive sign should be used instead. 

It is worth mentioning that the penalty method is well applicable to many problems; however, selecting appropriate penalty factors turns out to be problem-specific. If the penalty factors are too small, an infeasible may not get enough penalty. Thus, a feasible solution may be evolved in the optimization process. If too large values are used, a feasible solution can be of low quality and exploration over infeasible spaces is desirable. According to \cite{Yang2014Nature_Inspired}, the penalty factors $ \mu_{j}, \; j = \{1,\dots,2N+2\} $ are typically in the range of $ 10^{13} $ to $ 10^{15} $ for most problems. For experiments in this paper, we set all the penalty factors to be $ 10^{14} $ for the sake of simplicity. 

In Alg.~\ref{Alg:HHO}, the computation complexity of calculating the fitness values is $ \mathcal{O}(SD) $, where $ S $ is the population size and $ D $ is the dimension of a hawk. In the context of our problem in~\eqref{P1}, there are $ 2 $ coordinates $ \boldsymbol{w}=\{x_{u},y_{u}\} $ and $ N $ transmit power values corresponding to $ N $ GUs, thus $ D = 2 + N $. Updating the position vectors of all the hawks requires a computational complexity of $ \mathcal{O}(ND) $. Moreover, the computation of index functions for $ (2N+2) $ inequality constraints requires $ \mathcal{O}(S(2N+2)) $ times and the algorithm runs at most $ T $ iterations. As a results, in solving our problem formulated in~\eqref{P1} the HHO in Alg.~\ref{Alg:HHO} has the computational complexity level of $ \mathcal{O}(ST(D + 2N+2)) $. 

\begin{algorithm}[t]	
	\caption{Pseudo-code of the HHO algorithm}
	\label{Alg:HHO}
	\begin{algorithmic}[1]
		\State \textbf{Inputs}: The population dimensionality $ S $ and the maximum number of iterations $T$.
		\State \textbf{Outputs}: The position of the prey $ X_{p} $ and corresponding fitness value.
		\State Initialize the random population $ X_{i} \; (i=1,2,\ldots,S)$, and set the iteration index $ t = 0 $.
		
		\While{$ (t \leq T) $}	\label{Alg:T_iterations}
		\State Increase the iteration index $ t = t + 1 $ and calculate the \phantom{cci} fitness values of the hawks.
		\State \multiline{Select \textbf{$X_{p}$} with the highest fitness value as the position of the prey.} \label{Alg:bestPosition} 
		
		\For{$ (i = 1:S) $}  
			\State Update the initial energy as $ E_{0} = 2 \text{rand()} -1 $. 
			\State Update the jump strength as $ J = 2 (1 - \text{rand()}) $ and \phantom{cccccc} escaping energy $E$ via Eq.~\eqref{Eq:energyPrey}. 
			
			\If{ ($|E|\geq 1$)} \Comment{\texttt{Exploration phase}}
				\State Create a random number $ q $ in $ [0, 1] $.
				\If{($ q \geq 0.5$)}
					\State Update the position $ X_{i}(t+1) $ via Eq.\eqref{Ex:1stStrategy}. 
				\ElsIf{($ q > 0.5$)}
					\State Update the position $ X_{i}(t+1) $ via Eq.\eqref{Ex:2ndStrategy}. 
				\EndIf
			\EndIf
			
			\If {($|E|< 1$)} \Comment{\texttt{Exploitation phase}}
				\State Create a random probability $ r $.
				\If{($r \geq 0.5$ and $|E| \geq 0.5$ ) }  
					\State Update the position $ X_{i}(t+1) $ via Eq.~\eqref{Eq:SoftSiege}.
				\ElsIf{($r \geq 0.5$ and $|E|< 0.5$ ) } 
					\State Update the position $ X_{i}(t+1) $ via Eq.~\eqref{Eq:HardSiege}.
				\ElsIf{($r < 0.5$ and $|E|\geq 0.5$ ) } 
					\State Calculate the next move via~\eqref{Eq:SoftSiege_progressive_nm} and \phantom{ccccccccccccc} update the position $ X_{i}(t+1) $ via Eq.~\eqref{Eq:SoftSiege_progressive_pos}.
				\ElsIf{($r < 0.5$ and $|E|< 0.5$ ) }  
					\State Compute the next move via~\eqref{Eq:HardSiege_progressive_nm} and update \phantom{ccccccccccccc} the position $ X_{i}(t+1) $ via Eq.~\eqref{Eq:SoftSiege_progressive_pos}.
				\EndIf
			\EndIf
		\EndFor 
		
		\EndWhile
		
		\State \textbf{Return} The position of the prey $X_{p}$
	\end{algorithmic}
\end{algorithm}


\section{An HHO-based Optimizer for Artificial Neural Networks}
\label{Sec:HHO_ANN}
In this section, we propose an optimizer for ANNs based on the HHO algorithm and then apply it for joint power allocation and UAV's placement in UAV-assisted VLC systems using NOMA. The input to our proposed feedforward neural network (FNN) is only the normalized locations of all GUs and the outputs are normalized power allocation and UAV's placement vectors. After the training, the UAV's placement and transmit power can be computed in real-time via one-to-one mapping from the normalized locations of GUs using the trained FNN. 

\subsection{Feedforward Neural Networks for Joint Power Allocation and UAV's Placement}
FNNs are computational models consisting of many neurons (nodes), which are organized in layer-by-layer basis. The first and last layers are called the input layer and output layer, respectively, whereas ones between the input and output layers are called hidden layers. The term ``feedforward" in FNN means that the input being evaluated $ x $ goes through intermediate computations to finally reach the output $ y $, and there is no feedback connections from the output \cite{Goodfellow2016Deep}. Thanks to the structural representation, a single hidden layer FNN with a finite number of neurons has the capabilities to approximate any continuous function \cite{Hornik1991Approximation}.

\begin{figure}[t]
	\centering
	\includegraphics[width=0.90\linewidth]{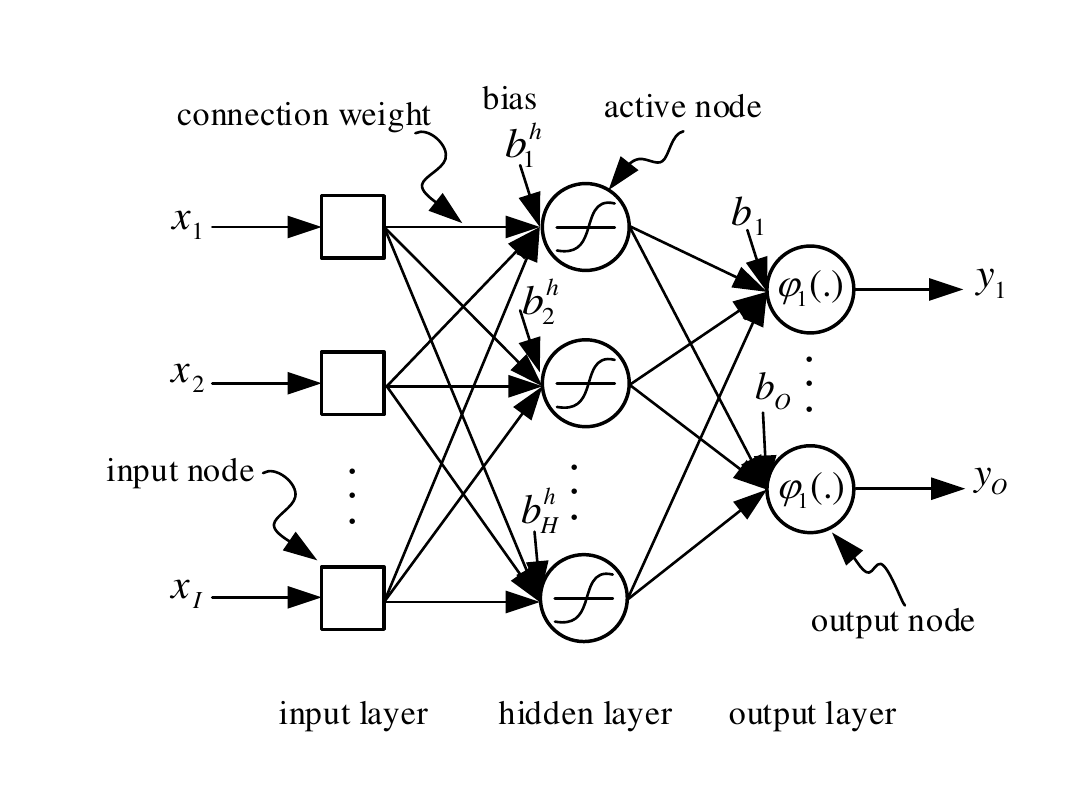}
	\caption{An FNN for joint power allocation and UAV's placement from a set of GUs' locations, where the input layer has $ I $ input nodes, the hidden layer has $ H $ activation functions, and the output layer has $ O $ nodes.}
	\label{Fig:ANN}	
\end{figure}

The illustration of a fully-connected FNN with only one single hidden layer for a VLC system is shown in Fig.~\ref{Fig:ANN}. Mathematically, the output $ y_{j} $ of a node (indicated as $ j $) in the hidden layer can be
computed as follows:
\begin{equation}\label{key}
z_{j} = \varphi_{i} \left( S_{j} \triangleq \sum\limits_{i = 1}^{I}w_{ij}x_{i} + b_{j} \right),
\end{equation}
where $ I $ is the number of input nodes, $ x_{i} $ is the $ i $-th input, $ w_{ij} $ is the connection weight from the $ i $-th node in the input layer to the $ j $-th node in the hidden layer, $ b_{j} $ is the bias of the $ j $-th hidden node, and $ \varphi_{j}(\cdot) $ is the activation function of the $ j $-th hidden node. As mentioned earlier, the optimization vector has the size of $ O = 2 + N $ {\color{black}(i.e., the number of neurons in the output layer is $ N+2 $)}, and the input is the locations of GUs, thus the input size is $ I = 2 \times N $. Since there is no standard method for selecting the number of neurons in the hidden layer, we adopt a rule-of-thumb method \cite{sheela2013review}, which advises that the number of neurons in the hidden layer should be between the numbers of neurons in the input and output layers. Additionally, the number of GUs supported simultaneously by NOMA is typically not less than two (i.e., $ N \geq 2 $), so we fix the number of neurons in the hidden layer as the size of the input layer, i.e., $ H = 2 + N $. 

Many activation functions can be used in the ANN, which is to limit the output amplitude of a neuron. Some commonly used activation functions are uni-polar sigmoid function, bipolar sigmoid function, hyperbolic tangent function, and rectified linear unit (ReLU) function. For example, the uni-polar sigmoid function can be expressed as \cite{Goodfellow2016Deep}
\begin{equation}\label{key}
z_{j} = \varphi_{j}(S_{j}) = \frac{1}{1 + \exp(-S_{j})}.
\end{equation}
After calculating the outputs of all neurons in the hidden layer, the output of the $ k $-th node 
can be calculated as follows:
\begin{equation}\label{key}
y_{k} = \sum\nolimits_{j = 1}^{H}w_{jk}z_{j} + b_{k}, \; k = \{1,\dots,O\}.
\end{equation}

It is worth mentioning that the input and output of the FNN are normalized values. More specifically, the normalized locations of GUs are computed as
\begin{equation}\label{key}
\hat{x}_{i} = \frac{ x_{i} - \E[x_{i}] }{ \sqrt{ \left( x_{i} - \E[x_{i}] \right)^{2} } } \text{ and } \hat{y}_{i} = \frac{ y_{i} - \E[y_{i}] }{ \sqrt{ \left( y_{i} - \E[y_{i}] \right)^{2} } },
\end{equation}
where $ x_{i} $ and $ y_{i} $ are $ x $-coordinate and $ y $-coordinate of GU $ i $, respectively. The outputs of the FNN are the normalized transmit power, denoted as $ \hat{\boldsymbol{p}} $, and the UAV's placement, denoted as $ \hat{\boldsymbol{w}} $. Then, the transmit power and UAV's placement can be calculated as $ \boldsymbol{p} = P_{\max}\hat{\boldsymbol{p}} $ and $ \boldsymbol{w} = R\hat{\boldsymbol{w}} $, respectively. 

\subsection{HHO for training FNNs}
Our purpose in this section is to utilize the HHO algorithm for training FNN so as to obtain the solution for joint power allocation and UAV's placement. For this purpose, two important points are how to encode the FNN optimization vector as the solution in the HHO algorithm and how to present the fitness function (also known as loss function in the FNN). Actually, several components of the FNN can be optimized, such as the connection weights, architecture (e.g., the number of hidden layers and the number of neurons in a hidden layer), training parameters, and node optimization (e.g., choosing the activation nodes) \cite{Ojha2017Metaheuristic}. Typically, scholars try to optimize the connection weights while fixing the other values. We follow this way to train the FNN for optimizing the problem formulated in this paper.

In the HHO algorithm, the position of each hawk needs to be encoded into a one-dimensional vector, which represents a candidate solution for the FNN. The vector includes three components: 1) the set of connection weights from the input layer to the hidden layer, 2) the set of weights connecting the hidden layer with the output layer, and 3) the set of biases. Thus, the size of each solution in the HHO is equivalent to the total number of weights and biases, i.e.,  $ D = I H + H O + H + O $. More specifically, the position of a hawk 
can be encoded as the following 
vector
\begin{align}
X = [ & \underbrace{w_{11}, \dots, w_{IH}}_{\text{input-hidden weights}}, \underbrace{w_{11}^{h}, \dots, w_{HO}^{h}}_{\text{hidden-output weights}}, \notag\\ & \underbrace{b_{1}^{h}, \dots, b_{H}^{h}}_{\text{hidden layer biases}}, \underbrace{b_{1}, \dots, b_{O}}_{\text{output layer biases}} ].
\end{align}
Here, $ w_{ij} $ denotes the weight connecting the $ i $-th node in the input layer with the $ j $-th node in the hidden layer, $ w_{jk}^{h} $ denotes the connection weight from the $ j $-th node in the hidden layer to the $ k $-th node in the output layer, and $ b_{j}^{h} $ ($ b_{k} $) denotes the bias associated with the $ j $-th ($ k $-th) node in the hidden layer (output layer).

Another important point is the selection of the fitness function, which is used to evaluate the positions of all the hawks. To this end, two following fitness functions, labeled as $ L_{T} $ and $ L_{L} $, are taken into account for optimizing the FNN:
\begin{align}
& L_{T} = - \sum\limits_{i \in \mathcal{N}}\log_{2}\left(1 + \frac{h_{i}p_{i}}{n_{0} + \sum\limits_{j \in \mathcal{N} : h_{j} > h_{i}}h_{i}p_{j}}\right), \label{Eq:lossFunction} \\
& L_{L} = \left( SR_{\text{HHOPAP}} - \sum\limits_{i \in \mathcal{N}}\log_{2}\left(1 + \frac{h_{i}p_{i}}{n_{0} + \sum\limits_{j : h_{j} > h_{i}}h_{i}p_{j}}\right) \right)^{2}, \label{Eq:lossFunction_L} 
\end{align}
where $ SR_{\text{HHOPAP}} $ is the sum rate achieved by the HHOPAP algorithm as in Alg.~\ref{Alg:HHO}. The first fitness function represents the minus of the total sum rate of all the GUs, i.e., the purpose is to train the FFN so that sum rate of all GUs, whereas the second fitness function denotes the squared difference between the sum rates obtained by the HHO trainer and HHOPAP algorithm, i.e., square deviation. 

We recall that for conventional algorithms (e.g., back propagation, Quickpro, Rprop, and conjugate gradient), information on the gradient of the loss function with respect to the optimization variables is required for training the FNN. For instance, a power allocation scheme was reported in \cite{Lee2019TransmitPower} to maximize the overall spectral efficiency (SEE) of underlay device-to-device (D2D) users, where the loss function is defined as the negative overall SE plus the violation in the interference threshold imposed to the cellular user. However, the gradient-based method may fall to local minima, has a slow convergence speed, and highly depends on initial parameters 
\cite{Aljarah2018Optimizing}.
{\color{black}Different from these algorithms, as a metaheuristic algorithm, the HHO relaxes the need for gradient information and offers high efficiency in avoiding trapping into local minima.} To illustrate the effectiveness of the proposed HHO-based algorithm, we will provide comparisons with three well-known metaheuristic algorithms: PSO, ES, and GA. 

\begin{algorithm}[t]	
	\caption{Pseudo-code of the HHOFNN trainer}
	\label{Alg:HHOFNN}
	\begin{algorithmic}[1]
		\State \textbf{Initialization}: Create a random population $ X_{i} \; (i=1,2,\ldots,S)$, initialize the stopping tolerance $ \epsilon $, and set the iteration index $ t = 1 $. 
		
		\While{$ | L_{T}(t) - L_{T}(t-1) | \geq \epsilon $}
		
		\State \multiline{The positions of the hawks are assigned to the weights and biases as potential FNNs.} \label{Alg2:St3}
		\State \multiline{Evaluate the fitness values corresponding to the FNNs. 
		The FNN with the lowest fitness value is then selected as the best solution.} \label{Alg2:St4}
		\State Update the positions of the hawks based on the HHO \phantom{11i} algorithm as shown in Alg.~\ref{Alg:HHO}. \label{Alg2:St5}
		
		\State \textcolor{black}{Set $ t = t+1 $.}
		
		\EndWhile
		
		\State \textbf{Return} the connection weights and biases of the FNN.
	\end{algorithmic}
\end{algorithm}

In summary, the HHO algorithm for training the FNN, named as HHOFNN, {\color{black}is described in Alg.~\ref{Alg:HHOFNN}. Step~\ref{Alg2:St3} is to assign the outputs of the HHOFNN algorithm to the vectors of power allocation and UAV's placement. Then, these vectors are used to calculate the fitness values via either~\eqref{Eq:lossFunction} or~\eqref{Eq:lossFunction_L}. In Step~\ref{Alg2:St4}, the FNN with the lowest fitness value is considered as the best FNN in the current iteration, which corresponds to Step~\eqref{Alg:bestPosition} in Alg.~\ref{Alg:HHO}. Step~\ref{Alg2:St5} refers to the main loop, lines 7-30, in Alg.~\ref{Alg:HHO}, i.e., this step is to update the FNNs, which are then used in the next iteration.} The algorithm is executed until the difference between loss values in two consecutive iterations is less than a stopping tolerance $ \epsilon $, i.e., the sum rate stays almost unchanged for two consecutive training times. As analyzed earlier, the computational complexity of Alg.~\ref{Alg:HHOFNN} is $ \mathcal{O}(SDT) $, where $ S $ is the size of hawks and $ T $ is the number of iterations needed to train the FNN.

\section{Simulation Results}
\label{Sec:Simulation}
In this section, we present numerical results to evaluate the performance of the proposed algorithm. For simulation settings, we consider a VLC system with a LED transmitter, which is mounted on a UAV, and 20 GUs (i.e., $ N = 20 $), which are randomly distributed in a coverage of 10 m $ \times $ 10 m $ \times $ 3 m. The maximal transmit power $ P_{\max} = 20 $ mW, the DC-offset $ A=20 $, the peak optical intensity $ B =30 $, and the noise power is $ n_{0} = -104 $ dBm. Moreover, the bandwidth $ B = 20 $~MHz, the disc radius $ R = 10 $ m, the physical area of the detector is $ 1 $ cm$ ^{2} $, the gain of the optical filter is $ 1.0 $, and $ \delta = 3\sqrt{5}/5 $ (for VLC systems with 4-ary PAM).

\begin{figure}[t]
	\centering
	\includegraphics[width=0.90\linewidth]{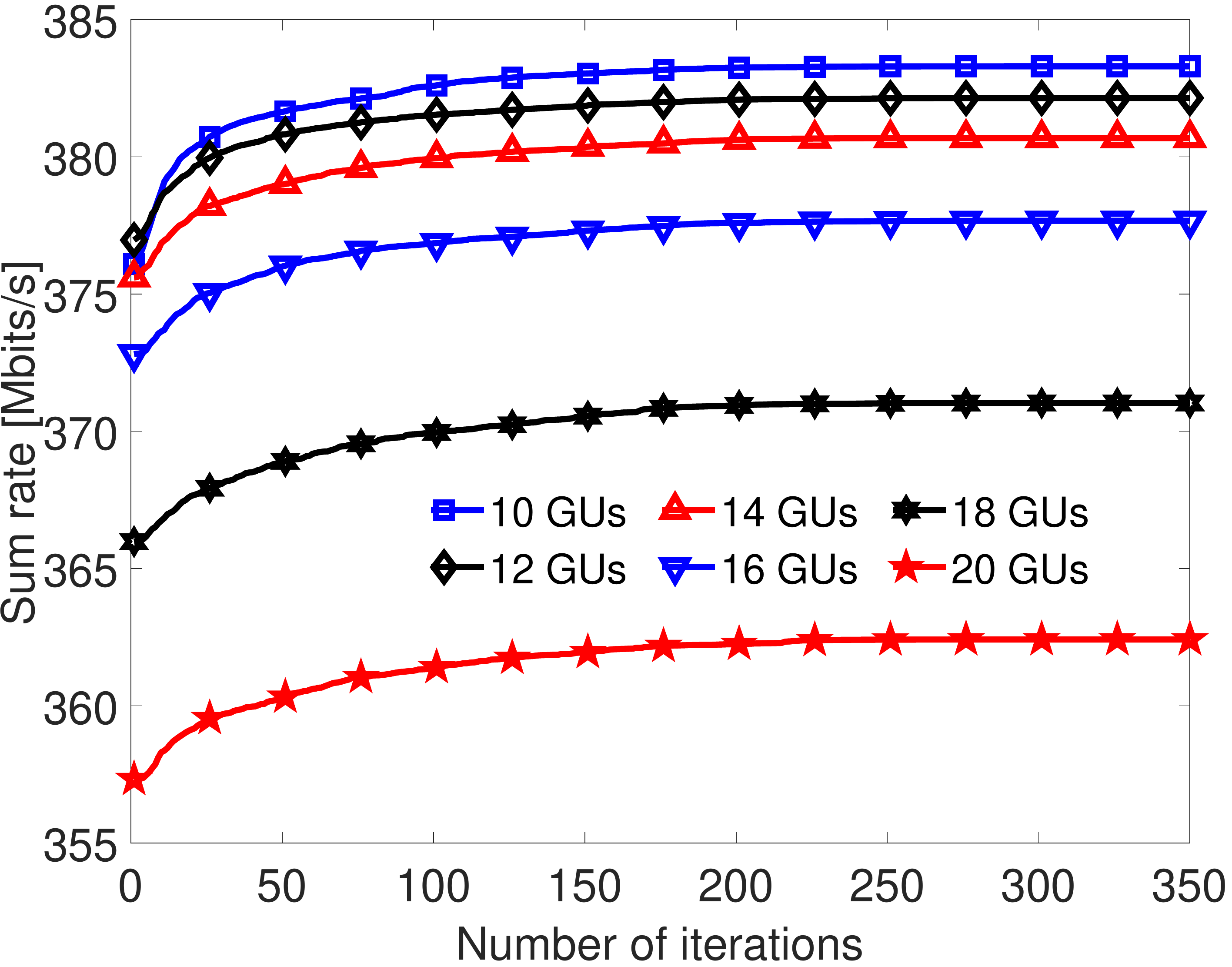}
	\caption{Convergence of the HHOPAP algorithm.}
	\label{Fig:convergence}	
\end{figure}

First, to investigate the convergence of the proposed algorithm, we set the maximum number of iterations to $ 350 $ (used to update the escaping energy of the prey, as in Eq.~\eqref{Eq:energyPrey}), the minimum required rate of each GU to $ 200 $ kbps, and the number of hawks to $ S = 30 $. As can be seen from Fig.~\ref{Fig:convergence}, the sum rate increases progressively for about $ 250 $ iterations and becomes constant for the remaining iterations. This shows that the proposed HHOPAP algorithm converges to the final solution within a reasonable time. Moreover, as the number of GUs increases, the sum rate decreases and experiences an increasing drop at the convergent solution. It is reasonable since GUs may be affected by more strong GUs (users with better channel conditions), thus even though SIC operation in NOMA works properly, GUs can receive much more interference from other strong GUs and the sum rate reduces accordingly. 

To show advantages of the proposed algorithm (labeled as HHOPAP), the following schemes are used for comparison:
\begin{itemize}
	\item \textit{GRPA}: this NOMA scheme was proposed in \cite{Marshoud2016NOMA_VLC}. The power allocation factor is set to be $ 0.4 $ and the UAV's placement is optimized by our algorithm. 
	\item \textit{RandP}: the UAV's position is randomly created within a disc as constrained in~\eqref{P1:g} and transmit power allocated to GUs is optimized by the HHO algorithm.
	\item \textit{Conventional OFDMA} \cite[Section 6.1.3]{tse2005fundamentals}: the bandwidth is equally divided among GUs. 
\end{itemize}

\begin{figure}[t]
	\centering
	\includegraphics[width=0.90\linewidth]{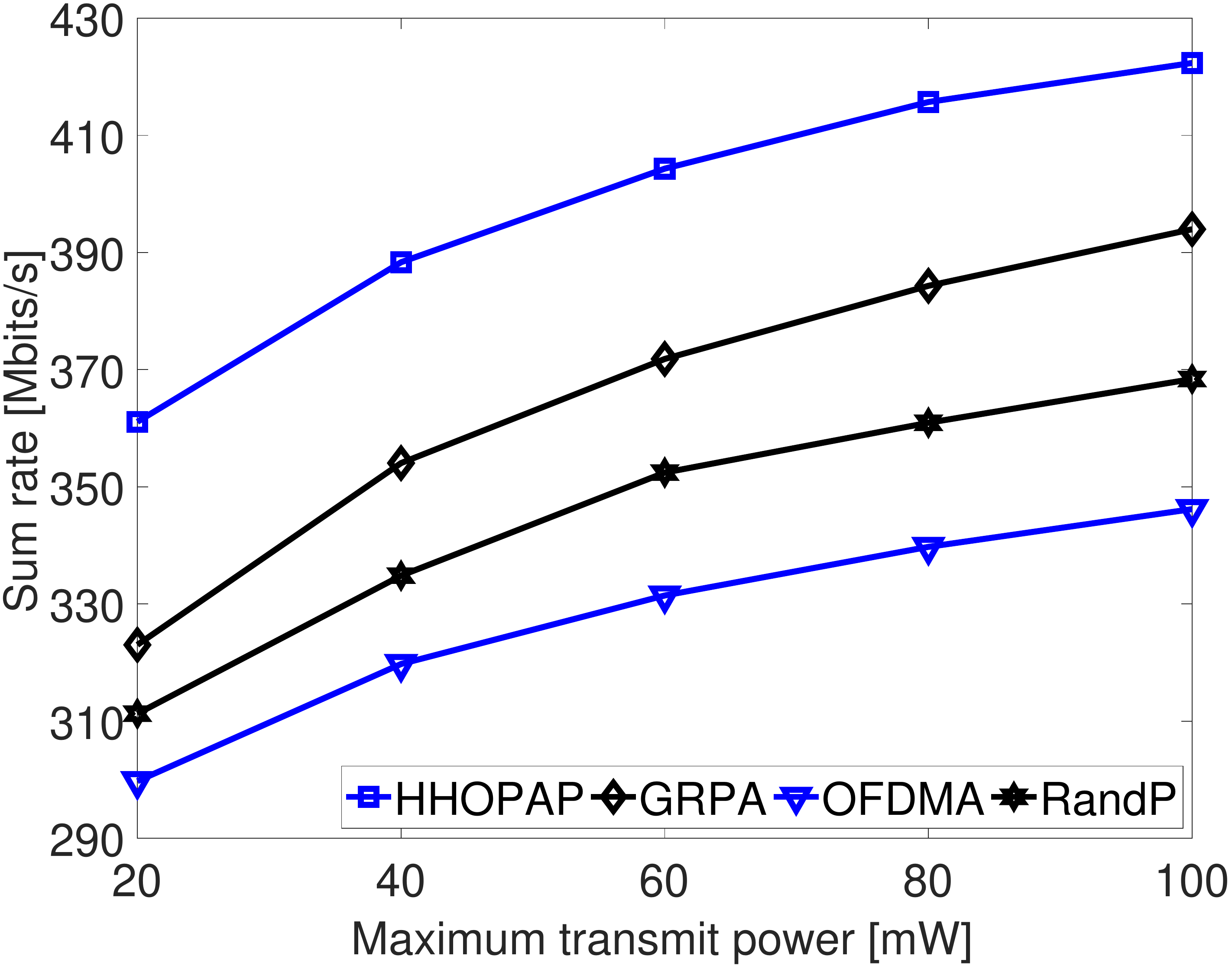}
	\caption{Sum rate against the maximum transmit power.}
	\label{Fig:script_vs_Pmax}	
\end{figure} 

Second, we evaluate the sum of data rates performance of the proposed HHOPAP and compare algorithms versus the maximum transmit power of the LED. Note that each plot below is obtained by averaging over 100 random network realizations, in which GUs are positioned randomly at each realization. It is observed from Fig.~\ref{Fig:script_vs_Pmax} that the sum rate experiences a steady increase as the maximum transmit power of the LED $ P_{\max} $ increases. It is because GUs can be allocated more power by the UAV and their achievable rate increases accordingly. However, the sum rate improves at a slower speed when the maximum transmit power increases. For instance, the proposed HHOPAP algorithm achieves an increase of $ 27.25 $ Mbits/s when the maximum transmit power grows from $ 20 $ to $ 40 $ mW, whereas the increase is only $ 11.38 $ ($ 6.67 $) Mbits/s for the values $ 60 $ and $ 80 $ mW ($ 80 $ and $ 100 $ mW). Another observation from the figure is that all the NOMA schemes (HHOPAP, GRPA, and RandP) outperform the conventional OFDMA scheme as they achieve higher sum rates. The figure also depicts that the proposed algorithm is better than alternative NOMA schemes (GRPA and RandP). This is because the GRPA considered a simple power allocation scheme, where the power is allocated to GUs based on the orders of their channel gains, and the RandP only optimizes the power allocated to GUs, whereas the UAV's placement is randomly selected. {\color{black}Above observations \emph{justify the importance of optimizing UAV's placement and NOMA power allocation} as NOMA schemes are superior to the OFDMA scheme, and the HHOPAP outperforms the GRPA (power is not optimized) and the RandP (UAV's placement is not optimized).}

\begin{figure}[t]
	\centering
	\includegraphics[width=0.90\linewidth]{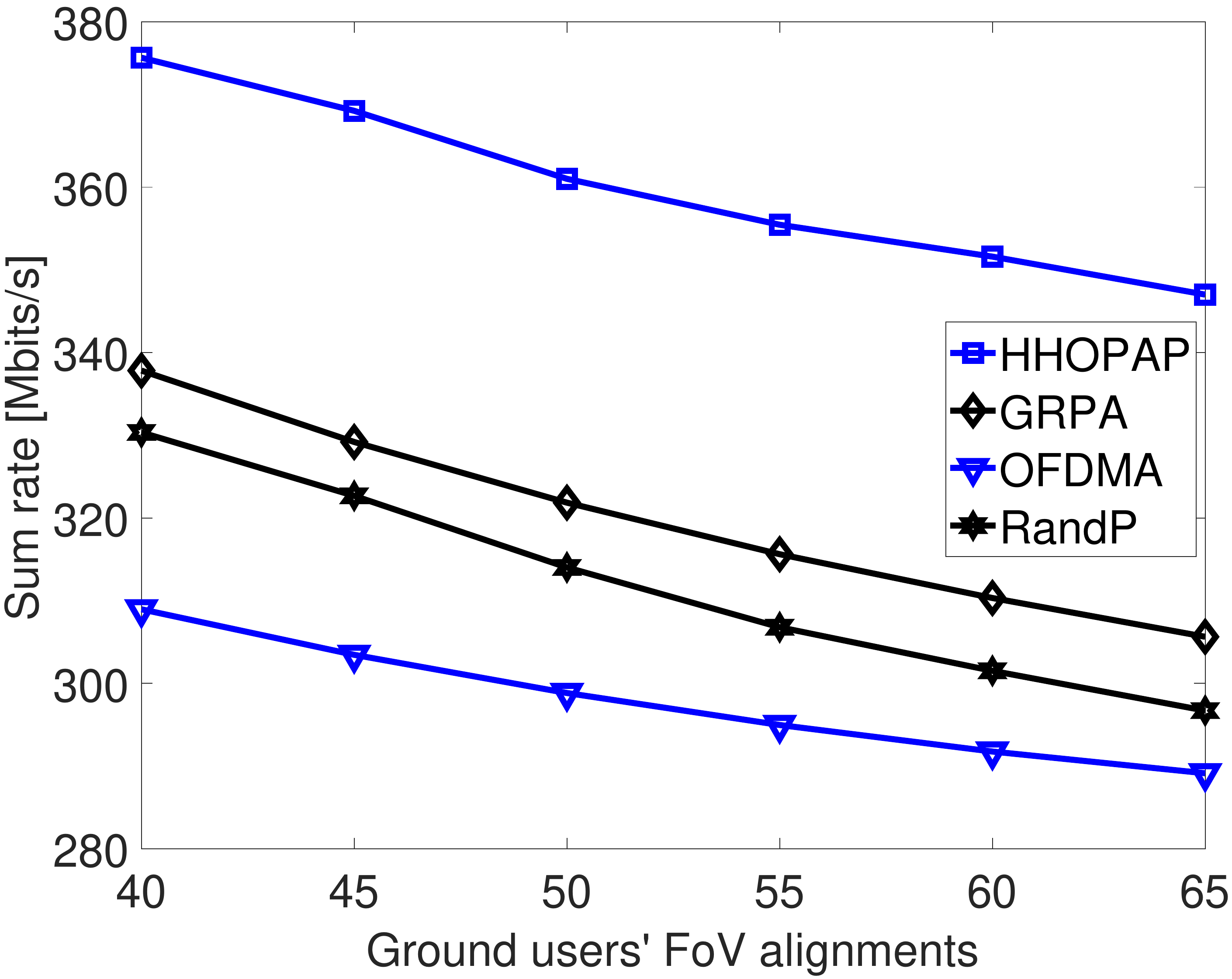}
	\caption{The sum-rate performance versus the GUs' FoV.}
	\label{Fig:script_vs_FoV}	
\end{figure} 

Next, Fig.~\ref{Fig:script_vs_FoV} shows the effect of the GU's FoV on the sum rate function. As can be seen from the figure, decreasing the GUs' FoV increases the sum rate, i.e., the FoVs of $ 40^{\circ} $ and $ 45^{\circ} $ achieve yield higher sum rate than the FoVs of $ 45^{\circ} $ and $ 50^{\circ} $, respectively. The reason for this is that the concentrator gain increases when the FoV $ \Phi $ varies from $ 0 $ to $ 90^{\circ} $, as justified in Eq.~\eqref{Eq:concentratorGain}, thus increasing the channel gain of GUs and the sum rate accordingly.
This result is highly interesting since if users, especially those who are using a smartphone, tend to keep the devices with their preferred direction. In that case, the FoV may be adjusted and the channel gain to the UAV-mounted LED may change significantly. Moreover, decreasing the FoV leads to increasing the coverage probability \cite{Obeed2019DC}. As a result, if the FoV is adjusted reasonably (i.e., decreasing the FoV's value while ensuring that the UAV is within the GUs' FoV), the sum rate can be enhanced.  Thanks to its high flexibility, the UAV  can be flexibly placed at a proper position, where it is under the FoV of all the GUs. 
Again, the NOMA schemes are superior to the conventional OFDMA scheme and the proposed algorithm achieves better sum-rate performance than all the alternative schemes.

\begin{figure}[t]
	\centering
	\includegraphics[width=0.90\linewidth]{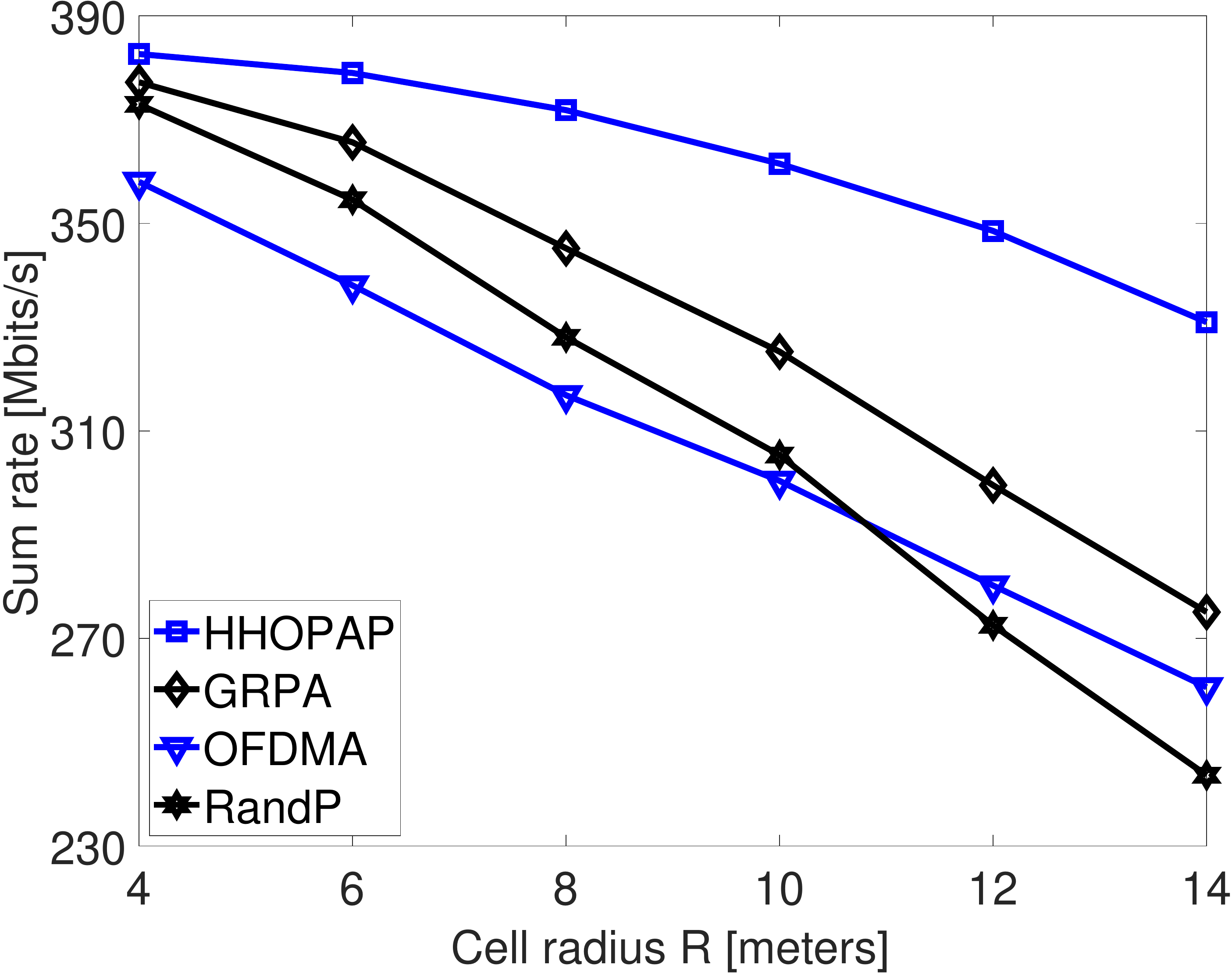}
	\caption{Comparison of the sum rate for different cell radius values.}
	\label{Fig:script_vs_discR}	
\end{figure}  

Fig.~\ref{Fig:script_vs_discR} studies the effect of the cell radius $ R $ on the sum-rate performance. The figure shows that increasing the cell radius leads to reducing the sum-rate performance. It is reasonable since for a given network setting, the larger the cell radius is, the longer the distance $ d_{i}, i \in \mathcal{N} $ from GUs to the UAV-mounted LED can be. Therefore, the channel gains between the UAV and GUs may become weaker (as shown in Eq.~\eqref{Eq:ChannelGain}), and the achievable rate reduces accordingly. What is more, the OFDMA scheme can surpass the RandP approach when the cell radius is sufficiently large (it is $ R = 12 $ meters for the simulation setting in this paper). This is because as the cell radius is larger, the UAV may be placed at a location very far from most GUs and thus the performance {\color{black}cannot be well compensated by NOMA power allocation in the RandP scheme. Again, this observation justifies the desirability of optimizing UAV's placement in UAV-assisted VLC systems.} Furthermore, the sum rate obtained by the proposed HHOPAP algorithm is the highest among all the alternative schemes. 

\begin{figure}[t]
	\centering
	\includegraphics[width=0.90\linewidth]{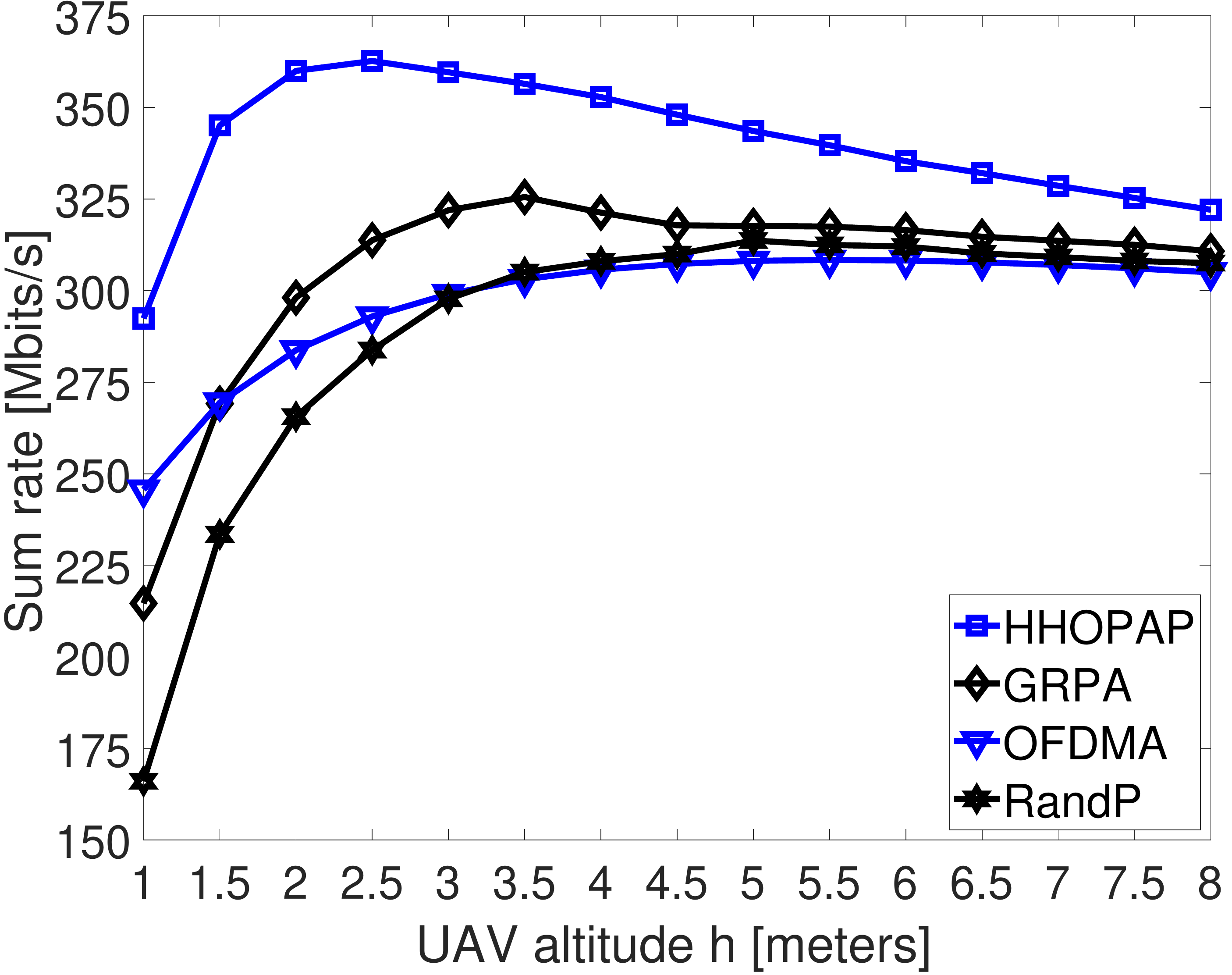}
	\caption{The sum-rate performance for different UAV's altitudes.}
	\label{Fig:script_vs_altitude}	
\end{figure}  

In Fig.~\ref{Fig:script_vs_altitude}, the sum rates achieved by different algorithms are compared versus the UAV's altitude (also known as hovering height in the literature). From the figure, it can be seen that the sum rate first increases when the UAV's altitude gets larger, but the sum rate starts to decrease when the UAV's altitude is large enough. This is due to the fact that increasing the UAV's altitude results in an increase in the channel gains. To illustrate this point, we consider an example, where the UAV and a GU $ i $ have the coordinates of $ (3,1,h) $ and $ (1,2,0) $, i.e., the distance is given as $ d_{i} = \sqrt{5 + h^{2}} $. When $ h = 1.5 $, $ d_{i} = \sqrt{6.25} $, $ \cos \phi_{i} = \cos \varphi_{i} = 1.5/\sqrt{6.25} $, and the channel gain is $ h_{i} = 1.5 \zeta / (6.25)^{3/2} $ (here $ \zeta $ is a constant and the value is computed via Eq.~\eqref{Eq:ChannelGain}), and similarly when $ h = 2.5 $, the channel gain is $ h_{i} = 2.5 \zeta / (11.25)^{3/2} $. Obviously, the altitude $ h = 2.5 $ offers a higher channel gain value than $ h = 1.5 $, and the GU's achievable rate increases as a consequence. 
However, the channel gain and sum-rate performance become worse if the altitude keeps increasing.  Using the same example as above, the channel gains for $ h = 4 $ and $ h = 5 $ are $ h_{i} = 4 \zeta / (29)^{3/2} $ and $ h_{i} = 5 \zeta / (38)^{3/2} $, respectively. 
Moreover, the OFDMA system is better than two NOMA schemes (GRPA and RandP) at lower values of the UAV altitude. The main reason is that the NOMA schemes typically achieve a better sum rate than OFDMA when the channel conditions of GUs are sufficiently distinctive, but this is not the case when the UAV altitude is small \cite{Ding2017aSurveyOnNOMA}. Besides, the figure shows that the proposed HHOPAP algorithm provides the best sum-rate performance in all cases of the altitude. 

\begin{figure*}[t]
	\centering
	\subfloat[MSE convergence curve for the Iris dataset. \label{Fig:Iris_Convergence}]{\includegraphics[width=0.45\linewidth]{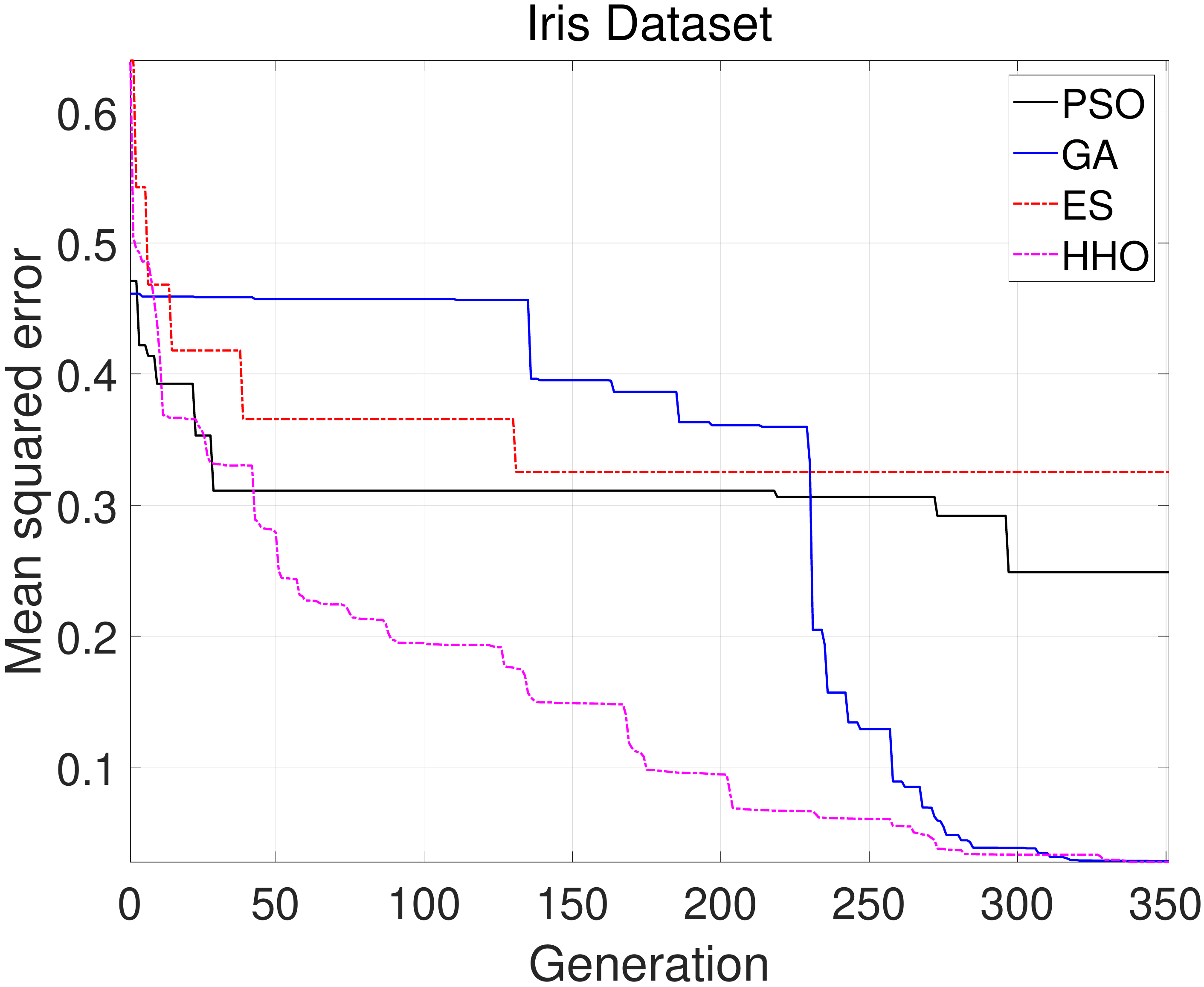}} \;
	\subfloat[MSE convergence curve for the Cancer dataset. \label{Fig:Cancer_Convergence}]{\includegraphics[width=0.46\linewidth]{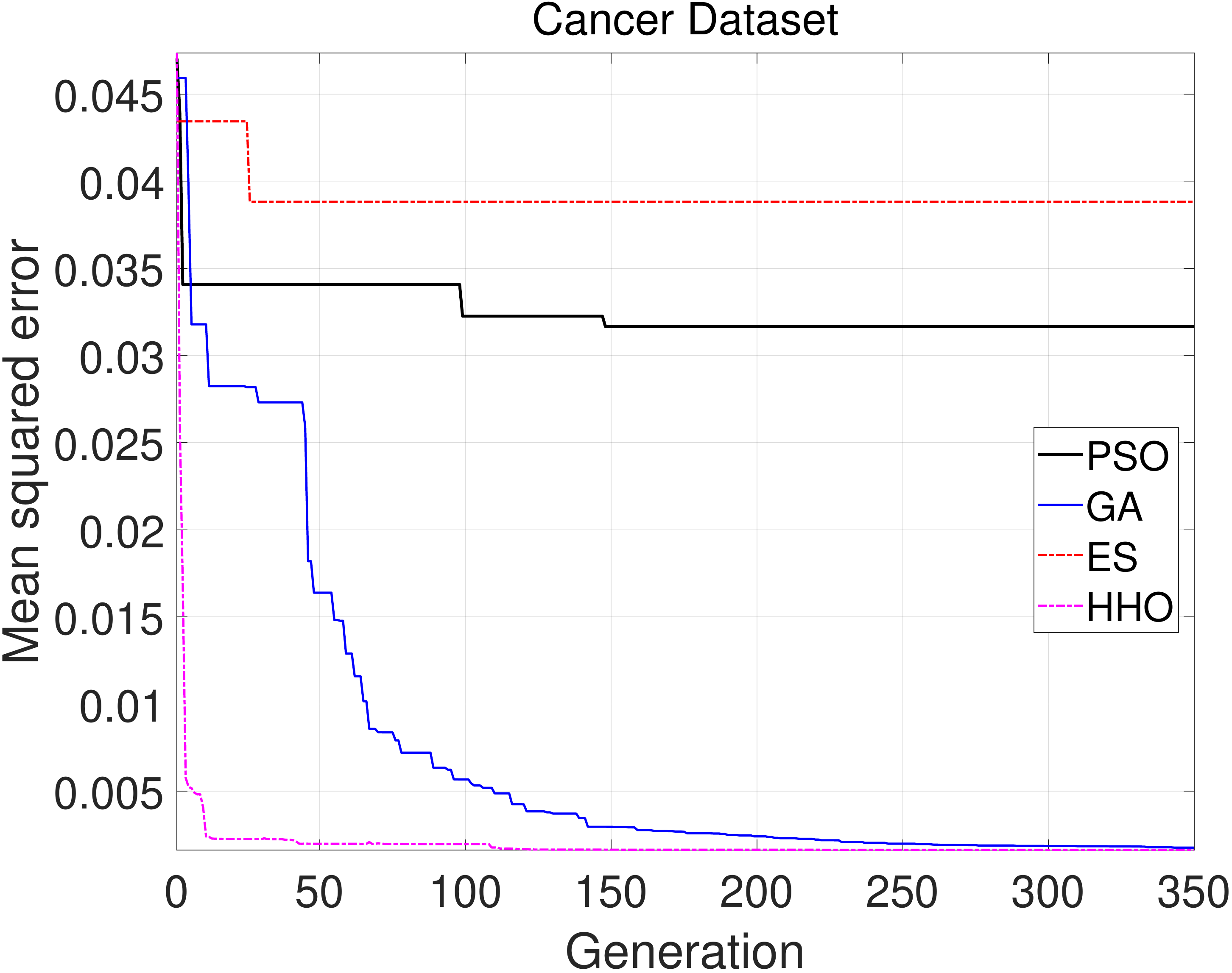}} \\
	\subfloat[Sum rate performance of different trainers. \label{Fig:sumrate_comparison_trainers}]{\includegraphics[width=0.46\linewidth]{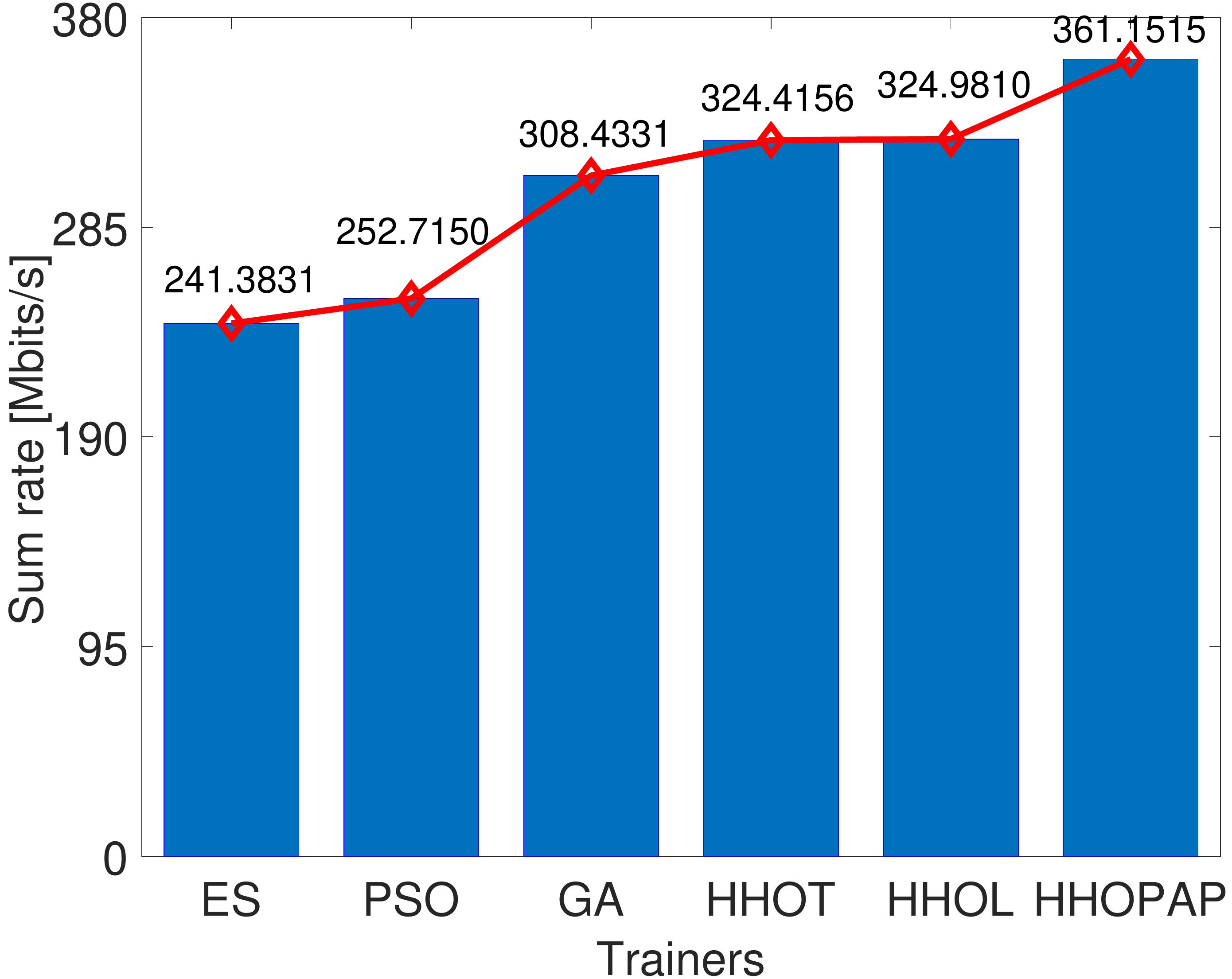}}
	\caption{Performance comparison of the HHO trainer.}
	\label{Fig:PerformanceComparison_vs_noSUs}
\end{figure*}

Finally, we show the performance of the HHO trainer and compare it with that of PSO, ES, and GA trainers. To verify the performance of the proposed HHO trainer on existing datasets, two popular ones (Iris and Cancer) are selected\footnote{Due to the space limitation, we only present two datasets, which are available at https://archive.ics.uci.edu/ml/datasets.php.}. From Figs.~\ref{Fig:Iris_Convergence} and~\ref{Fig:Cancer_Convergence}, the HHO achieves the lowest mean square error (MSE), which are $ 0.027562 $ and $ 0.0016149 $ for the Iris and Cancer datasets, respectively. This result {\color{black}\emph{demonstrates the capability of the HHO in avoiding falling into the local minima trap}}, which usually happens in conventional trainers like PSO and ES.
As can be observed from Fig.~\ref{Fig:sumrate_comparison_trainers}, the HHO trainer achieves almost the same sum-rate performance with two types of fitness functions. In particular, the first kind (labeled as HHOT) can achieve the sum rate of $ 34.4156 $ Mbits/s, whereas the second one (labeled as HHOL) achieves the sum rate of $ 324.9810 $ Mbits/s. Compared with the other trainers (ES, PSO, and GA), the HHO offers a better performance in terms of the sum rate. Another observation is that the sum rate obtained by the HHO trainer is closest to the proposed HHOPAP with only about 10\% loss, which shows that the HHO trainer is a promising solution for real-time applications.

\section{Conclusion and Future Works} 
\label{Sec:Conclusion}
In this paper, we have studied the integration of UAV and NOMA into VLC systems. We have investigated the sum-rate maximization problem, which was solved by facilitating the HHO algorithm. In addition, we have utilized the HHO for training an FNN so that the UAV's placement and power allocation can be computed in real-time. Various simulation settings and performance metrics have been tested to evaluate the effectiveness and superiority of the proposed HHO algorithm and HHOFNN trainer. {\color{black}Furthermore, we have observed an obvious gain of jointly optimizing UAV's placement and NOMA power allocation, when compared with the schemes of optimizing either UAV's placement (i.e., GRPA) or power allocation (i.e., RandP).}


{\color{black}Many issues and open problems stemmed from this paper should be investigated in the future. Firstly, since the performance loss is currently about 10\%, it is a valuable contribution to further improve the performance of the HHO trainer. Secondly, when fixed-wing UAVs are considered, the joint three-dimensional trajectory optimization and power allocation problem is an important issue. Thirdly, because of the limited on-board energy of UAVs, the energy-efficient design of UAV-assisted VLC is desirable to study. Next, the LoS connection between the UAV and GUs may pose physical layer security issues, which demand more efforts in the future. Besides, it is interesting to investigate an algorithm, which can find the globally optimal solution for the original problem~\eqref{P1}. Finally, it is desirable to examine the HHO algorithm with other ANN architectures and apply the HHO algorithm to other optimization problems of resource allocation in wireless and communications.}


%
%




\begin{thebibliography}{10}
\providecommand{\url}[1]{#1}
\csname url@samestyle\endcsname
\providecommand{\newblock}{\relax}
\providecommand{\bibinfo}[2]{#2}
\providecommand{\BIBentrySTDinterwordspacing}{\spaceskip=0pt\relax}
\providecommand{\BIBentryALTinterwordstretchfactor}{4}
\providecommand{\BIBentryALTinterwordspacing}{\spaceskip=\fontdimen2\font plus
\BIBentryALTinterwordstretchfactor\fontdimen3\font minus
  \fontdimen4\font\relax}
\providecommand{\BIBforeignlanguage}[2]{{%
\expandafter\ifx\csname l@#1\endcsname\relax
\typeout{** WARNING: IEEEtran.bst: No hyphenation pattern has been}%
\typeout{** loaded for the language `#1'. Using the pattern for}%
\typeout{** the default language instead.}%
\else
\language=\csname l@#1\endcsname
\fi
#2}}
\providecommand{\BIBdecl}{\relax}
\BIBdecl

\bibitem{Marshoud2016NOMA_VLC}
H.~{Marshoud}, V.~M. {Kapinas}, G.~K. {Karagiannidis}, and S.~{Muhaidat},
  ``Non-orthogonal multiple access for visible light communications,''
  \emph{IEEE Photonics Technology Letters}, vol.~28, no.~1, pp. 51--54, Jan.
  2016.

\bibitem{Ding2017aSurveyOnNOMA}
Z.~Ding, X.~Lei, G.~K. Karagiannidis, R.~Schober, J.~Yuan, and V.~K. Bhargava,
  ``A survey on non-orthogonal multiple access for {5G} networks: Research
  challenges and future trends,'' \emph{IEEE Journal on Selected Areas in
  Communications}, vol.~35, no.~10, pp. 2181--2195, Oct. 2017.

\bibitem{Pham2019ASurvey_MEC}
\BIBentryALTinterwordspacing
Q.-V. Pham, F.~Fang, V.~N. Ha, M.~Le, Z.~Ding, L.~B. Le, and W.-J. Hwang, ``A
  survey of multi-access edge computing in {5G} and beyond: Fundamentals,
  technology integration, and state-of-the-art,'' \emph{CoRR}, 2019. [Online].
  Available: \url{arxiv.org/abs/1906.08452}
\BIBentrySTDinterwordspacing

\bibitem{Nguyen2010Matlab}
H.~Q. {Nguyen}, J.~H.-{Choi}, M.~{Kang}, Z.~{Ghassemlooy}, D.~H. {Kim},
  S.~K.-Lim, T.~G.-{Kang}, and C.~G. {Lee}, ``A {MATLAB}-based simulation
  program for indoor visible light communication system,'' in \emph{2010 7th
  International Symposium on Communication Systems, Networks Digital Signal
  Processing (CSNDSP 2010)}, Newcastle upon Tyne, UK, Jul. 2010, pp. 537--541.

\bibitem{Pham2019CoalitionalGames}
Q.-V. Pham, T.~H. {Nguyen}, Z.~{Han}, and W.-J. Hwang, ``Coalitional games for
  computation offloading in {NOMA}-enabled multi-access edge computing,''
  \emph{IEEE Transactions on Vehicular Technology}, vol.~69, no.~2, pp.
  1982--1993, Feb. 2020.

\bibitem{Yang2017FairNOMA}
Z.~{Yang}, W.~{Xu}, and Y.~{Li}, ``Fair non-orthogonal multiple access for
  visible light communication downlinks,'' \emph{IEEE Wireless Communications
  Letters}, vol.~6, no.~1, pp. 66--69, Feb. 2017.

\bibitem{Yin2016Performance}
L.~{Yin}, W.~O. {Popoola}, X.~{Wu}, and H.~{Haas}, ``Performance evaluation of
  non-orthogonal multiple access in visible light communication,'' \emph{IEEE
  Transactions on Communications}, vol.~64, no.~12, pp. 5162--5175, Dec. 2016.

\bibitem{Zhang2017UserGrouping}
X.~{Zhang}, Q.~{Gao}, C.~{Gong}, and Z.~{Xu}, ``User grouping and power
  allocation for {NOMA} visible light communication multi-cell networks,''
  \emph{IEEE Communications Letters}, vol.~21, no.~4, pp. 777--780, Apr. 2017.

\bibitem{Lin2019OpticalPD}
B.~{Lin}, X.~{Tang}, and Z.~{Ghassemlooy}, ``Optical power domain {NOMA} for
  visible light communications,'' \emph{IEEE Wireless Communications Letters},
  vol.~8, no.~4, pp. 1260--1263, Aug. 2019.

\bibitem{Nasir2019UAV_Enabled}
A.~A. {Nasir}, H.~D. {Tuan}, T.~Q. {Duong}, and H.~V. {Poor}, ``{UAV}-enabled
  communication using {NOMA},'' \emph{IEEE Transactions on Communications},
  vol.~67, no.~7, pp. 5126--5138, Jul. 2019.

\bibitem{Mu2019UplinkNOMA}
\BIBentryALTinterwordspacing
X.~Mu, Y.~Liu, L.~Guo, and J.~Lin, ``Uplink non-orthogonal multiple access for
  {UAV} communications,'' \emph{CoRR}, 2019. [Online]. Available:
  \url{arxiv.org/abs/1906.06523}
\BIBentrySTDinterwordspacing

\bibitem{Seo2019UplinkNOMA}
J.~{Seo}, S.~{Pack}, and H.~{Jin}, ``Uplink {NOMA} random access for
  {UAV}-assisted communications,'' \emph{IEEE Transactions on Vehicular
  Technology}, vol.~68, no.~8, pp. 8289--8293, Aug. 2019.

\bibitem{Liu2019Placement}
X.~{Liu}, J.~{Wang}, N.~{Zhao}, Y.~{Chen}, S.~{Zhang}, Z.~{Ding}, and F.~R.
  {Yu}, ``Placement and power allocation for {NOMA}-{UAV} networks,''
  \emph{IEEE Wireless Communications Letters}, vol.~8, no.~3, pp. 965--968,
  Jun. 2019.

\bibitem{Sohail2019EnergyEfficient}
M.~F. {Sohail}, C.~Y. {Leow}, and S.~{Won}, ``Energy-efficient non-orthogonal
  multiple access for {UAV} communication system,'' \emph{IEEE Transactions on
  Vehicular Technology}, vol.~68, no.~11, pp. 10\,834--10\,845, Nov. 2019.

\bibitem{Zhao2019JointTrajectory}
N.~{Zhao}, X.~{Pang}, Z.~{Li}, Y.~{Chen}, F.~{Li}, Z.~{Ding}, and M.~{Alouini},
  ``Joint trajectory and precoding optimization for {UAV}-assisted {NOMA}
  networks,'' \emph{IEEE Transactions on Communications}, vol.~67, no.~5, pp.
  3723--3735, May 2019.

\bibitem{Pang2019UplinkPrecoding}
X.~{Pang}, G.~{Gui}, N.~{Zhao}, W.~{Zhang}, Y.~{Chen}, Z.~{Ding}, and
  F.~{Adachi}, ``Uplink precoding optimization for {NOMA} cellular-connected
  {UAV} networks,'' \emph{IEEE Transactions on Communications}, vol.~68, no.~2,
  pp. 1271--1283, Feb 2020.

\bibitem{Yang2019PowerEfficient}
Y.~{Yang}, M.~{Chen}, C.~{Guo}, C.~{Feng}, and W.~{Saad}, ``Power efficient
  visible light communication ({VLC}) with unmanned aerial vehicles ({UAVs}),''
  \emph{IEEE Communications Letters}, vol.~23, no.~7, pp. 1272--1275, Jul.
  2019.

\bibitem{Wang2019DeepLearningUAV}
Y.~Wang, M.~Chen, Z.~Yang, T.~Luo, and W.~Saad, ``Deep learning for optimal
  deployment of {UAVs} with visible light communications,'' \emph{arXiv
  preprint arXiv:1912.00752}, 2019.

\bibitem{chen2019joint}
M.~Chen, Z.~Yang, W.~Saad, C.~Yin, H.~V. Poor, and S.~Cui, ``A joint learning
  and communications framework for federated learning over wireless networks,''
  \emph{arXiv preprint arXiv:1909.07972}, 2019.

\bibitem{yang2019energy}
Z.~Yang, M.~Chen, W.~Saad, C.~S. Hong, and M.~Shikh-Bahaei, ``Energy efficient
  federated learning over wireless communication networks,'' \emph{arXiv
  preprint arXiv:1911.02417}, 2019.

\bibitem{luong2019applications}
N.~C. Luong, D.~T. Hoang, S.~Gong, D.~Niyato, P.~Wang, Y.-C. Liang, and D.~I.
  Kim, ``Applications of deep reinforcement learning in communications and
  networking: A survey,'' \emph{IEEE Communications Surveys \& Tutorials},
  vol.~21, no.~4, pp. 3133--3174, 2019.

\bibitem{Heidari2019HHO}
A.~A. Heidari, S.~Mirjalili, H.~Faris, I.~Aljarah, M.~Mafarja, and H.~Chen,
  ``Harris hawks optimization: Algorithm and applications,'' \emph{Future
  Generation Computer Systems}, vol.~97, pp. 849--872, 2019.

\bibitem{golilarz2019new}
N.~A. Golilarz, A.~Addeh, H.~Gao, L.~Ali, A.~M. Roshandeh, H.~M. Munir, and
  R.~U. Khan, ``A new automatic method for control chart patterns recognition
  based on {ConvNet} and harris hawks meta heuristic optimization algorithm,''
  \emph{IEEE Access}, vol.~7, pp. 149\,398--149\,405, Dec. 2019.

\bibitem{abbasi2019application}
A.~Abbasi, B.~Firouzi, and P.~Sendur, ``On the application of harris hawks
  optimization ({HHO}) algorithm to the design of microchannel heat sinks,''
  \emph{Engineering with Computers}, pp. 1--20, Dec. 2019.

\bibitem{Girmay2019JointChannel}
G.~G. {Girmay}, Q.-V. Pham, and W.-J. Hwang, ``Joint channel and power
  allocation for device-to-device communication on licensed and unlicensed
  band,'' \emph{IEEE Access}, vol.~7, no. Feb., pp. 22\,196--22\,205, 2019.

\bibitem{Pham2020Whale}
Q.-V. Pham, S.~{Mirjalili}, N.~{Kumar}, M.~{Alazab}, and W.-J. Hwang, ``Whale
  optimization algorithm with applications to resource allocation in wireless
  networks,'' \emph{IEEE Transactions on Vehicular Technology}, 2020, in press.

\bibitem{Huang2019Wireless}
J.~{Huang}, Y.~{Zhou}, Z.~{Ning}, and H.~{Gharavi}, ``Wireless power transfer
  and energy harvesting: Current status and future prospects,'' \emph{IEEE
  Wireless Communications}, vol.~26, no.~4, pp. 163--169, Aug. 2019.

\bibitem{Deng2018Twinkle}
H.~Deng, J.~Li, A.~Sayegh, S.~Birolini, and S.~Andreani, ``Twinkle: A flying
  lighting companion for urban safety,'' in \emph{Proceedings of the Twelfth
  International Conference on Tangible, Embedded, and Embodied Interaction},
  ser. TEI '18.\hskip 1em plus 0.5em minus 0.4em\relax Stockholm, Sweden: ACM,
  2018, pp. 567--573.

\bibitem{Nurzhan2019Unmanned}
N.~Kalikulov, R.~C. Kizilirmak, and M.~Uysal,
  ``{Unmanned-aerial-vehicle-assisted cooperative communications for visible
  light communications-based vehicular networks},'' \emph{Optical Engineering},
  vol.~58, no.~8, pp. 1--9, Aug. 2019.

\bibitem{Kahn1997WirelessIC}
J.~M. {Kahn} and J.~R. {Barry}, ``Wireless infrared communications,''
  \emph{Proceedings of the IEEE}, vol.~85, no.~2, pp. 265--298, Feb. 1997.

\bibitem{Pham2018AlphaFairness}
Q.-V. Pham and W.-J. Hwang, ``\BIBforeignlanguage{English}{$\alpha$-fair
  resource allocation in non-orthogonal multiple access systems},''
  \emph{\BIBforeignlanguage{English}{IET Communications}}, vol.~12, no.~2, pp.
  179--183, Jan. 2018.

\bibitem{Ali2016Dynamic}
M.~S. {Ali}, H.~{Tabassum}, and E.~{Hossain}, ``Dynamic user clustering and
  power allocation for uplink and downlink non-orthogonal multiple access
  ({NOMA}) systems,'' \emph{IEEE Access}, vol.~4, pp. 6325--6343, Aug. 2016.

\bibitem{Pham2017Fairness}
Q.-V. Pham and W.-J. Hwang, ``Fairness-aware spectral and energy efficiency in
  spectrum-sharing wireless networks,'' \emph{IEEE Transactions on Vehicular
  Technology}, vol.~66, no.~11, pp. 10\,207--10\,219, Nov. 2017.

\bibitem{dugatkin1997cooperation}
L.~A. Dugatkin, \emph{Cooperation among animals: an evolutionary
  perspective}.\hskip 1em plus 0.5em minus 0.4em\relax Oxford University Press
  on Demand, 1997.

\bibitem{Ling2017LevyFlight}
Y.~{Ling}, Y.~{Zhou}, and Q.~{Luo}, ``Lévy flight trajectory-based whale
  optimization algorithm for global optimization,'' \emph{IEEE Access}, vol.~5,
  pp. 6168--6186, Apr. 2017.

\bibitem{Xie2019ImprovedBHO}
W.~{Xie}, J.~S. {Wang}, and Y.~{Tao}, ``Improved black hole algorithm based on
  golden sine operator and levy flight operator,'' \emph{IEEE Access}, vol.~7,
  pp. 161\,459--161\,486, Nov. 2019.

\bibitem{Jordehi2015Review}
A.~R. Jordehi, ``A review on constraint handling strategies in particle swarm
  optimisation,'' \emph{Neural Computing and Applications}, vol.~26, no.~6, pp.
  1265--1275, Aug. 2015.

\bibitem{Coello2002Theoretical}
C.~A.~C. Coello, ``Theoretical and numerical constraint-handling techniques
  used with evolutionary algorithms: a survey of the state of the art,''
  \emph{Computer Methods in Applied Mechanics and Engineering}, vol. 191,
  no.~11, pp. 1245--1287, Jan. 2002.

\bibitem{Yang2014Nature_Inspired}
X.-S. Yang, \emph{Nature-Inspired Optimization Algorithms}.\hskip 1em plus
  0.5em minus 0.4em\relax 32 Jamestown Road, London NW1 7BY: Elsevier, 2014.

\bibitem{Goodfellow2016Deep}
I.~Goodfellow, Y.~Bengio, and A.~Courville, \emph{Deep learning}.\hskip 1em
  plus 0.5em minus 0.4em\relax MIT press, 2016.

\bibitem{Hornik1991Approximation}
K.~Hornik, ``Approximation capabilities of multilayer feedforward networks,''
  \emph{Neural Networks}, vol.~4, no.~2, pp. 251--257, 1991.

\bibitem{sheela2013review}
K.~G. Sheela and S.~N. Deepa, ``Review on methods to fix number of hidden
  neurons in neural networks,'' \emph{Mathematical Problems in Engineering},
  vol. 2013, 2013.

\bibitem{Ojha2017Metaheuristic}
V.~K. Ojha, A.~Abraham, and V.~Sn{\'a}šel, ``Metaheuristic design of
  feedforward neural networks: A review of two decades of research,''
  \emph{Engineering Applications of Artificial Intelligence}, vol.~60, pp.
  97--116, Apr. 2017.

\bibitem{Lee2019TransmitPower}
W.~{Lee}, M.~{Kim}, and D.~{Cho}, ``Transmit power control using deep neural
  network for underlay device-to-device communication,'' \emph{IEEE Wireless
  Communications Letters}, vol.~8, no.~1, pp. 141--144, Feb. 2019.

\bibitem{Aljarah2018Optimizing}
I.~Aljarah, H.~Faris, and S.~Mirjalili, ``Optimizing connection weights in
  neural networks using the whale optimization algorithm,'' \emph{Soft
  Computing}, vol.~22, no.~1, pp. 1--15, Jan. 2018.

\bibitem{tse2005fundamentals}
D.~Tse and P.~Viswanath, \emph{Fundamentals of wireless communication}.\hskip
  1em plus 0.5em minus 0.4em\relax Cambridge university press, 2005.

\bibitem{Obeed2019DC}
M.~{Obeed}, H.~{Dahrouj}, A.~M. {Salhab}, S.~A. {Zummo}, and M.~{Alouini},
  ``{DC}-bias and power allocation in cooperative {VLC} networks for joint
  information and energy transfer,'' \emph{IEEE Transactions on Wireless
  Communications}, vol.~18, no.~12, pp. 5486--5499, Dec. 2019.

\end{thebibliography}
%
\balance
\end{document}